\documentclass[11pt,twoside]{article} 

\RequirePackage[OT1]{fontenc}
\RequirePackage{amsthm,amsmath,amsfonts,amssymb}
\RequirePackage{epsfig, graphicx}
\RequirePackage{latexsym}
\RequirePackage{psfrag}
\usepackage{color}

\usepackage{fullpage}
\usepackage{epsf,epsfig,subfigure}
\usepackage{fancyheadings}
\usepackage{graphics,graphicx}
\usepackage{caption}
\usepackage{enumerate}

\usepackage{epsfig}

\theoremstyle{plain}

\newtheorem{theo}{Theorem}[section]

\newtheorem{lem}{Lemma}[section]
\newtheorem{prop}{Proposition}[section]
\newtheorem{cor}{Corollary}[section]

\theoremstyle{definition} 

\newtheorem{nota}{Notation}[section]
\newtheorem{de}{Definition}[section]
\newtheorem{exa}{Example}[section]
\newtheorem{as}{Assumption}[section]
\newtheorem{alg}{Algorithm}[section]

\newcommand{\btheo}{\begin{theo}}
\newcommand{\bde}{\begin{de}}
\newcommand{\ble}{\begin{lem}}
\newcommand{\bpr}{\begin{prop}}
\newcommand{\bno}{\begin{nota}}
\newcommand{\bex}{\begin{exa}}
\newcommand{\bcor}{\begin{cor}}
\newcommand{\spro}{\begin{proof}}
\newcommand{\bas}{\begin{as}}
\newcommand{\balg}{\begin{alg}}

\newcommand{\etheo}{\end{theo}}
\newcommand{\ede}{\end{de}}
\newcommand{\ele}{\end{lem}}
\newcommand{\epr}{\end{prop}}
\newcommand{\eno}{\end{nota}}
\newcommand{\eex}{\end{exa}}
\newcommand{\ecor}{\end{cor}}
\newcommand{\fpro}{\end{proof}}
\newcommand{\eas}{\end{as}}
\newcommand{\ealg}{\end{alg}}

\theoremstyle{plain}

\newtheorem{theos}{Theorem}
\newtheorem{props}{Proposition}
\newtheorem{lems}{Lemma}
\newtheorem{cors}{Corollary}

\theoremstyle{definition}
\newtheorem{exas}{Example}
\newtheorem{algs}{Algorithm}
\newtheorem{asss}{Asumption}
\newtheorem{defns}{Definition}

\newcommand{\btheos}{\begin{theos}}
\newcommand{\etheos}{\end{theos}}
\newcommand{\bprops}{\begin{props}}
\newcommand{\eprops}{\end{props}}
\newcommand{\bdes}{\begin{defns}}
\newcommand{\edes}{\end{defns}}
\newcommand{\blems}{\begin{lems}}
\newcommand{\elems}{\end{lems}}
\newcommand{\bcors}{\begin{cors}}
\newcommand{\ecors}{\end{cors}}
\newcommand{\bexs}{\begin{exas}}
\newcommand{\eexs}{\end{exas}}
\newcommand{\balgs}{\begin{algs}}
\newcommand{\ealgs}{\end{algs}}
\newcommand{\bass}{\begin{asss}}
\newcommand{\eass}{\end{asss}}

\newcommand{\noise}{w}
\newcommand{\noiseRot}{\tilde{w}}

\newcommand{\Grad}{\nabla}

\newcommand{\Bias}{\ensuremath{S}}
\newcommand{\Par}{\eta}
\newcommand{\PredPar}{\gamma}

\newcommand{\PenPar}{\nu}
\newcommand{\PenParHat}{\ensuremath{\widehat{\PenPar}}}

\newcommand{\Step}[1]{\ensuremath{\alpha^{#1}}}

\newcommand{\diag}{\mbox{diag}}

\newcommand{\real}{\ensuremath{\mathbb{R}}}

\newcommand{\numobs}{\ensuremath{n}}
\newcommand{\inprod}[2]{\ensuremath{\langle #1 , \, #2 \rangle}}
\newcommand{\EmpRisk}{\ensuremath{\mathcal{L}}}

\newcommand{\mprob}{\ensuremath{\mathbb{P}}}
\newcommand{\defn}{\ensuremath{: \, = }}

\newcommand{\y}{\ensuremath{y}}

\newcommand{\yi}{\ensuremath{\y_i}}
\newcommand{\exi}{\ensuremath{x_i}}

\newcommand{\wi}{\ensuremath{w_i}}

\newcommand{\LTP}{\ensuremath{L^2(\mathbb{P})}}

\newcommand{\LTPn}{\ensuremath{{L^2(\mathbb{P}_n)}}}

\newcommand{\Eig}{\ensuremath{\lambda}}

\newcommand{\Hil}{\ensuremath{\mathcal{H}}}
\newcommand{\Xspace}{\ensuremath{\mathcal{X}}}

\newcommand{\fstar}{\ensuremath{f^*}}

\newcommand{\order}{\ensuremath{\mathcal{O}}}

\newcommand{\gencon}{\ensuremath{c}}
\newcommand{\plaincon}{\ensuremath{c}}

\newcommand{\Rad}{\ensuremath{\mathcal{R}}}

\newcommand{\Ball}{\ensuremath{\mathbb{B}}}

\newcommand{\Exs}{\ensuremath{\mathbb{E}}}

\newcommand{\fhat}{\ensuremath{\widehat{f}}}

\newcommand{\Ker}{\ensuremath{\mathbb{K}}}

\newcommand{\kerrank}{\ensuremath{m}}

\newcommand{\Xcal}{\ensuremath{\mathcal{X}}}

\newcommand{\LocGauss}{\ensuremath{\mathcal{Q}_{\numobs}}}

\newcommand{\LocGaussEmp}{\ensuremath{\widehat{\mathcal{Q}}_{\numobs}}}

\newcommand{\Rvar}{\ensuremath{Z_\numobs}}
\newcommand{\RvarHat}{\ensuremath{\widehat{Z}_\numobs}}

\newcommand{\smooth}{\ensuremath{\nu}}

\newcommand{\EmpKer}{\ensuremath{K}}

\newcommand{\Nat}{\ensuremath{\mathbb{N}}}

\newcommand{\FUN}[1]{\ensuremath{f^{#1}(x_1^\numobs)}}
\newcommand{\Ydata}{\ensuremath{y_1^\numobs}}

\newcommand{\Wdata}{\ensuremath{w}}
\newcommand{\BVEC}[1]{\ensuremath{\theta^{#1}}}

\newcommand{\RUNSUM}[1]{\ensuremath{\eta_{#1}}}

\newcommand{\EmpKerRoot}{\ensuremath{\sqrt{\EmpKer}}}

\newcommand{\origweight}{\ensuremath{\omega}}
\newcommand{\iter}{\ensuremath{t}}

\newcommand{\EmpRiskTrans}{\ensuremath{\widetilde{\EmpRisk}}}

\newcommand{\STEP}[1]{\ensuremath{\Step{#1}}}

\newcommand{\Dmat}{\ensuremath{\Lambda}} 

\newcommand{\kereigemp}{\ensuremath{\widehat{\kereig}}}
\newcommand{\kereig}{\ensuremath{\lambda}}

\newcommand{\SHRINK}[1]{\ensuremath{\Bias^{#1}}}
\newcommand{\SHRINKSQ}[1]{\ensuremath{(\Bias^{#1})^2}}

\newcommand{\STOP}{\ensuremath{\widehat{T}}}

\newcommand{\RISKHO}{\ensuremath{R_{\mbox{\tiny{HO}}}}}
\newcommand{\RISKSURE}{\ensuremath{R_{\mbox{\tiny{SU}}}}}
\newcommand{\RISKOracle}{\ensuremath{R_{\mbox{\tiny{OR}}}}}
\newcommand{\STOPHO}{\ensuremath{\STOP_{\mbox{\tiny{HO}}}}}
\newcommand{\STOPSURE}{\ensuremath{\STOP_{\mbox{\tiny{SU}}}}}
\newcommand{\STOPOracle}{\ensuremath{\STOP_{\mbox{\tiny{OR}}}}}

\newcommand{\Strain}{\ensuremath{S_{\mbox{\tiny{tr}}}}}

\newcommand{\ftrain}{\ensuremath{f_{\mbox{\tiny{tr}}}}}

\newcommand{\FUNIT}[1]{\ensuremath{f_{#1}}}

\newcommand{\FUNSTAR}{\ensuremath{f^*(x_1^\numobs)}}

\newcommand{\BIASSQ}{\ensuremath{B_\iter^2}}
\newcommand{\MYVAR}{\ensuremath{V_\iter}}

\newcommand{\widgraph}[2]{\includegraphics[keepaspectratio,width=#1]{#2}}

\newcommand{\DiagOpt}{\ensuremath{D}}

\newcommand{\trace}{\ensuremath{\operatorname{trace}}}

\newcommand{\EPAR}{\ensuremath{\varepsilon}}
\newcommand{\EMPCRIT}{\ensuremath{\widehat{\varepsilon}_\numobs}}
\newcommand{\POPCRIT}{\ensuremath{\varepsilon_\numobs}}

\newcommand{\RadEmp}{\ensuremath{\widehat{\Rad}}}

\newcommand{\MYPRED}[1]{\ensuremath{\PredPar^{#1}}}

\newcommand{\TAILPAR}{\ensuremath{\delta}}

\newcommand{\matsnorm}[2]{|\!|\!| #1 | \! | \!|_{{#2}}}

\newcommand{\fronorm}[1]{\ensuremath{\matsnorm{#1}{\scriptsize{\mbox{F}}}}}
\newcommand{\opnorm}[1]{\ensuremath{\matsnorm{#1}{\scriptsize{\mbox{op}}}}}

\newcommand{\KERSHRINK}{\ensuremath{R^\PenPar}}

\newcommand{\fbou}{\ensuremath{B}}

\long\def\comment#1{}

\makeatletter
\def\@cite#1#2{[\if@tempswa #2 \fi #1]}
\makeatother

\makeatletter
\long\def\@makecaption#1#2{
        \vskip 0.8ex
        \setbox\@tempboxa\hbox{\small {\bf #1:} #2}
        \parindent 1.5em  
        \dimen0=\hsize
        \advance\dimen0 by -3em
        \ifdim \wd\@tempboxa >\dimen0
                \hbox to \hsize{
                        \parindent 0em
                        \hfil 
                        \parbox{\dimen0}{\def\baselinestretch{0.96}\small
                                {\bf #1.} #2
                                } 
                        \hfil}
        \else \hbox to \hsize{\hfil \box\@tempboxa \hfil}
        \fi
        }
\makeatother

\begin{document}
	\begin{center}
	{\bf{\LARGE{Early stopping and non-parametric regression:
              \\ An optimal data-dependent stopping rule}}}

	\vspace*{.2in}

	\begin{tabular}{cc}
	Garvesh Raskutti$^\dagger$ & Martin J. Wainwright$^{\dagger,\ast}$ \\
	\end{tabular}
	\begin{tabular}{c}
	 Bin Yu$^{\dagger,\ast}$
	\end{tabular}

	\vspace*{.2in}

	\begin{tabular}{c}
	Department of Statistics$^\dagger$, and Department of
        EECS$^\ast$ \\
	UC Berkeley, Berkeley, CA 94720
	\end{tabular}

	\end{center}

\vspace*{.1in}

\begin{abstract}
The strategy of early stopping is a regularization technique based on
choosing a stopping time for an iterative algorithm.  Focusing on
non-parametric regression in a reproducing kernel Hilbert space, we
analyze the early stopping strategy for a form of gradient-descent
applied to the least-squares loss function.  We propose a
data-dependent stopping rule that does not involve hold-out or
cross-validation data, and we prove upper bounds on the squared error
of the resulting function estimate, measured in either the
$L^2(\mprob)$ and $L^2(\mprob_\numobs)$ norm.  These upper bounds lead
to minimax-optimal rates for various kernel classes, including Sobolev
smoothness classes and other forms of reproducing kernel Hilbert
spaces.  We show through simulation that our stopping rule compares
favorably to two other stopping rules, one based on hold-out data and the
other based on Stein's unbiased risk estimate.  We also establish a
tight connection between our early stopping strategy and the solution
path of a kernel ridge regression estimator.

\end{abstract}


\section{Introduction}

The phenomenon of overfitting is ubiquitous throughout statistics.  It
is especially problematic in nonparametric problems, where some form
of regularization is essential to prevent overfitting. In the problem
of nonparametric regression, the most classical form of regularization
is that of Tikhonov regularization, where a quadratic smoothness
penalty is added to the least-squares loss. An alternative and
algorithmic approach to regularization is based on early stopping of
an iterative algorithm, such as gradient descent applied to the
unregularized loss function. The main advantage of early stopping for
regularization, as compared to penalized forms, is lower computational
complexity.

The idea of early stopping has a fairly lengthy history, dating back
to the 1970's in the context of the Landweber iteration.  (For
instance, see the paper by Strand~\cite{Strand74}, with follow-up work
by Anderssen and Prenter~\cite{AndrerssenPrenter81} as well as
Wahba~\cite{Wahba87}.) Early stopping has also been widely used in
neural networks (e.g.,~\cite{MorganBourlard90}), in which stochastic
gradient descent is used to estimate the network parameters. Past work
provided intuitive arguments for the benefits of early stopping. It
was argued that each step of an iterative algorithm will reduce bias
but increase variance, so early stopping ensures the variance of the
estimator is not too high. However, prior to the 1990s, there had been
little theoretical justification for these claims.  A more recent line
of work has provided theoretical justification for various types of
early stopping, including boosting algorithms
(e.g.,~\cite{BartlettTraskin07, BuhlmannYu03, Freund97, Jiang04,
  Mason99, Yao05,ZhangYu05}), greedy methods~\cite{Barron08}, gradient
descent over reproducing kernel Hilbert spaces
(e.g.~\cite{Bauer07,Caponnetto06,CaponnettoYao06,DeVito10, Yao05}),
the conjugate gradient algorithm~\cite{BlanchKram10}, and the power
method for eigenvalue computation~\cite{Orecchia11}. Most relevant to
our work is the work of B\"{u}hlmann and Yu~\cite{BuhlmannYu03}, who
derived optimal mean-squared error bounds for $L^2$-boosting with
early stopping in the case of fixed design regression.  However, these
optimal rates are based on an ``oracle'' stopping rule, one that
cannot be computed based on the data.  Thus, their work left open the
following natural question: is there a data-dependent and easily
computable stopping rule that produces a minimax-optimal estimator?

The main contribution of this paper is to answer this question in the
affirmative for a certain class of non-parametric regression problems,
in which the underlying regression function belongs to a reproducing
kernel Hilbert space (RKHS).  In this setting, a standard estimator is
the method of kernel ridge regression (e.g.,~\cite{Wahba}), which
minimizes a weighted sum of the least-squares loss with a squared
Hilbert norm penalty as a regularizer.  Instead of a penalized form of
regression, we analyze early stopping of an iterative update that is
equivalent to gradient descent on the least-squares loss in an
appropriately chosen coordinate system.  By analyzing the mean-squared
error of our iterative update, we derive a data-dependent stopping
rule that provides the optimal trade-off between the estimated bias
and variance at each iteration. In particular, our stopping rule is
based on the first time that a running sum of step-sizes after $t$
steps increases above the critical trade-off between bias and
variance.  For Sobolev spaces and other types of kernel classes, we
show that the function estimate obtained by this stopping rule
achieves minimax-optimal estimation rates in both the empirical and
population norms.  Importantly, our stopping rule does not require the
use of cross-validation or hold-out data.

In more detail, our first main result (Theorem~\ref{ThmMain}) provides
bounds on the squared prediction error for all iterates prior to the
stopping time, and a lower bound on the squared error for all
iterations after the stopping time.  These bounds are applicable to
the case of fixed design, where as our second main result
(Theorem~\ref{ThmRandDesign}) provides similar types of upper bounds
for randomly sampled covariates.  These bounds are stated in terms of
the squared $\LTP$ norm, as opposed to the prediction error or $\LTPn$
(semi)norm defined by the data.  Both of these theorems apply to any
reproducing kernel, and lead to specific predictions for different
kernel classes, depending on their eigendecay.  For the case of low
rank kernel classes and Sobolev spaces, we prove that our stopping
rule yields a function estimate that achieves the minimax optimal rate
(up to a constant pre-factor), so that the bounds from our analysis
are essentially unimprovable. Our proof is based on a combination of
analytic techniques~\cite{BuhlmannYu03} with techniques from empirical
process theory~\cite{vandeGeer}.  We complement these theoretical
results with simulation studies that compare its performance to other
rules, in particular a method using hold-out data to estimate the
risk, as well as a second method based on Stein's Unbiased Risk
Estimate (SURE).  In our experiments for first-order Sobolev kernels,
we find that our stopping rule performs favorably compared to these
alternatives, especially as the sample size grows.  Finally, in
Section~\ref{SecRidgeCompare}, we provide an explicit link between our
early stopping strategy and the kernel ridge regression estimator.


\section{Background and problem formulation}
\label{SecProbSetup}

We begin by introducing some background on non-parametric regression
and reproducing kernel Hilbert spaces, before turning to a precise
formulation of the problem studied in this paper.


\subsection{Non-parametric regression and kernel classes}
\label{SecRKHS}

Suppose that our goal is to use a covariate $X \in \Xspace$ to predict
a real-valued response $Y \in \real$.  We do so by using a function
$f: \Xspace \rightarrow \real$, where the value $f(x)$ represents our
prediction of $Y$ based on the realization $X = x$. In terms of
mean-squared error, the optimal choice is the \emph{regression
  function} defined by \mbox{$\fstar(x) \defn \Exs[Y \mid x]$.} In the
problem of non-parametric regression with random design, we observe
$\numobs$ samples of the form \mbox{$\{ (\exi, \yi), i = 1, \ldots,
  \numobs \}$,} each drawn independently from some joint distribution
on the Cartesian product $\Xspace \times \real$, and our goal is to
estimate the regression function $\fstar$.  Equivalently, we observe
samples of the form
\begin{align}
\label{EqnLinObs}
\yi & = \fstar(\exi) + \wi, \quad \mbox{for $i = 1,2, \ldots,
  \numobs$,}
\end{align}
where $\wi \defn \yi - \fstar(\exi)$ is a zero-mean noise random variable.
Throughout this paper, we assume that the random variables $\wi$ are
\emph{sub-Gaussian} with parameter $\sigma$, meaning that
\begin{align}
\label{EqnDefnSubGaussian}
\Exs[e^{t w_i}] & \leq e^{t^2 \sigma^2/2} \quad \mbox{for all $t \in
  \real$.}
\end{align}
For instance, this sub-Gaussian condition is satisfied for normal
variates $w_i \sim N(0, \sigma^2)$, but it also holds for various
non-Gaussian random variables.  Parts of our analysis also apply to
the fixed design setting, in which we condition on a particular
realization $\{x_i\}_{i=1}^\numobs$ of the covariates.

In order to estimate
the regression function, we make use of the machinery of reproducing
kernel Hilbert spaces~\cite{Aron50,Wahba,Gu01}.  Using $\mprob$ to
denote the marginal distribution of the covariates, we consider a
Hilbert space \mbox{$\Hil \subset \LTP$,} meaning a family of
functions $g: \Xcal \rightarrow \real$, with $\|g\|_{\LTP} < \infty$,
and an associated inner product $\inprod{\cdot}{\cdot}_\Hil$ under
which $\Hil$ is complete. The space $\Hil$ is a reproducing kernel
Hilbert space (RKHS) if there exists a symmetric function $\Ker: \Xcal
\times \Xcal \rightarrow \real_+$ such that: (a) for each $x
\in \Xspace$, the function $\Ker(\cdot, x)$ belongs to the Hilbert
space $\Hil$, and (b) we have the reproducing relation $f(x) =
\inprod{f}{\Ker(\cdot, x)}_{\Hil}$ for all $f \in \Hil$.  Any such
kernel function must be positive semidefinite; under suitable
regularity conditions, Mercer's theorem~\cite{Mercer09} guarantees
that the kernel has an eigen-expansion of the form
\begin{align}
\label{EqnMercer}
\Ker(x, x') & = \sum_{k=1}^\infty \kereig_k \phi_k(x)
\phi_k(x'),
\end{align}
where $\lambda_1 \geq \lambda_2 \geq \lambda_3 \geq \ldots \geq 0$ are a non-negative sequence of eigenvalues, and $\{\phi_k\}_{k=1}^\infty$
are the associated eigenfunctions, taken to be orthonormal in $\LTP$.
The decay rate of the eigenvalues will play a crucial role in our
analysis.

Since the eigenfunctions $\{\phi_k\}_{k=1}^\infty$ form an orthonormal
basis, any function $f \in \Hil$ has an expansion of the form $f(x) =
\sum_{k=1}^{\infty} \sqrt{\kereig_k} a_{k} \phi_k(x)$, where 
for all $k$ such that $\kereig_k > 0$, the coefficients
\begin{align*}
a_k & \defn \frac{1}{\sqrt{\kereig_k}} \inprod{f}{\phi_k}_{\LTP} =
\int_\Xspace f(x) \phi_k(x) \, d\, \mprob(x)
\end{align*}
are rescaled versions of the generalized Fourier
coefficients. Associated with any two functions in $\Hil$---where
\mbox{$f = \sum_{k = 1}^\infty \sqrt{\kereig_k} a_k \phi_k$} and
\mbox{$g = \sum_{k =1}^\infty \sqrt{\kereig_k} b_k \phi_k$}---are two
distinct inner products. The first is the usual inner product in the
space $\LTP$---namely, \mbox{$\inprod{f}{g}_{\LTP} \defn \int_\Xspace
  f(x) g(x) \, d\, \mprob(x)$.} By Parseval's theorem, it has an
equivalent representation in terms of the expansion coefficients and
kernel eigenvalues---that is,
\begin{align*}
\inprod{f}{g}_{\LTP} & = \sum_{k=1}^\infty \kereig_k a_k b_k.
\end{align*}
The second inner product, denoted $\inprod{f}{g}_{\Hil}$, is the one
that defines the Hilbert space; it can be written in terms of the
expansion coefficients as
\begin{align*}
\inprod{f}{g}_\Hil & = \sum_{k=1}^\infty {a_k
  b_k}.
\end{align*}
Using this definition, the Hilbert ball of radius $1$ for the Hilbert
space $\Hil$ with eigenvalues $\{\kereig_k\}_{k=1}^\infty$ and
eigenfunctions $\{\phi_k\}_{k=1}^\infty$ takes the form
\begin{align}
\Ball_\Hil(1) & \defn \big \{ f = \sum_{k=1}^{\infty} \sqrt{\kereig_k}
b_{k} \phi_k \quad \mbox{for some} \quad \sum_{k=1}^{\infty} {b_k^2}
\leq 1 \big\}.
\end{align}
The class of reproducing kernel Hilbert spaces contains many
interesting classes that are widely used in practice, including
polynomials of degree $d$, Sobolev spaces with smoothness $\smooth$,
and Gaussian kernels. For more background and examples on reproducing
kernel Hilbert spaces, we refer the reader to various standard
references~\cite{Aron50,Saitoh88,Scholkopf02,Wahba,Weinert82}.

Throughout this paper, we assume that any function $f$ in the unit
ball of the Hilbert space is uniformly bounded, meaning that there is
some constant $\fbou < \infty$ such that
\begin{align}
\label{EqnCond}
\|f\|_{\infty} & \defn \sup_{x \in \Xspace} |f(x)| \leq \fbou \qquad
\mbox{for all $f \in \Ball_\Hil(1)$.}
\end{align}
This boundedness condition~\eqref{EqnCond} is satisfied for any RKHS
with a kernel such that $\sup_{x \in \Xspace} \Ker(x,x) \leq \fbou$.
Kernels of this type include the Gaussian and Laplacian kernels, the
kernels underlying Sobolev and other spline classes, as well as as
well as any trace class kernel with trignometric eigenfunctions.  The
boundedness condition~\eqref{EqnCond} is quite standard in
non-asymptotic analysis of non-parametric regression
procedures~\cite[e.g.]{vandeGeer}.


\subsection{Gradient update equation}
\label{SecGradStep}

We now turn to the form of the gradient update that we study in this
paper. Given the samples \mbox{$\{(x_i, y_i)\}_{i=1}^\numobs$},
consider minimizing the least-squares loss function
\begin{align}
\EmpRisk(f) & \defn \frac{1}{2 \numobs} \sum_{i=1}^\numobs {\big (\yi
  - f(x_i) \big)^2}
\end{align}
over some subset of the Hilbert space $\Hil$.  By the representer
theorem~\cite{Kimeldorf71}, it suffices to restrict attention to
functions $f$ belonging to the span of the kernel functions $\{
\Ker(\cdot, x_i), i = 1, \ldots, \numobs \}$.  Accordingly, we adopt
the parameterization
\begin{align}
\label{EqnRepresenter}
f(\cdot) & = \frac{1}{\sqrt{n}} \sum_{i=1}^{n}{\origweight_i
  \Ker(\cdot ,x_i)},
\end{align}
for some coefficient vector $\origweight \in \real^\numobs$.  Here the
rescaling by $1/\sqrt{\numobs}$ is for later theoretical convenience.

Our gradient descent procedure is based on a parameterization of the
least-squares loss that involves the \emph{empirical kernel matrix}
\mbox{$\EmpKer \in \mathbb{R}^{n \times n}$} with entries
\begin{equation}
\label{EqnDefnEmpiricalKernel}
[\EmpKer]_{ij} = \frac{1}{n}\Ker(x_i, x_j) \qquad \mbox{for $i, j = 1,
  2, \ldots, \numobs$.}
\end{equation}
For any positive semidefinite kernel function, this matrix must be
positive semidefinite, and so has a unique symmetric square root
denoted by $\EmpKerRoot$.  Introducing the convenient shorthand
\mbox{$\Ydata \defn \big(y_1 \; y_2 \; \cdots y_\numobs \big) \in
  \real^\numobs$,} we can then write the least-squares loss in the
form
\begin{align*}
\EmpRisk(\origweight) & = \frac{1}{2 \numobs} \| \Ydata -
\sqrt{\numobs} \EmpKer \origweight \|_2^2.
\end{align*}
A direct approach would be to perform gradient descent on this form of
the least-squares loss.  For our purposes, it turns out to be more
natural to perform gradient descent in the transformed co-ordinate
system $\BVEC{} = \EmpKerRoot \, \origweight$.  Some straightforward
calculations (see Appendix~\ref{AppGrad} for details) yield that the
gradient descent algorithm in this new co-ordinate system generates a
sequence of vectors $\{\BVEC{\iter}\}_{\iter=0}^\infty$ via the
recursion
\begin{align}
\label{EqnGradBvec}
\BVEC{\iter+1} & = \BVEC{\iter} - \Step{\iter} \Big( \EmpKer \,
\BVEC{\iter} - \frac{1}{\sqrt{\numobs}} \EmpKerRoot \, \Ydata \Big),
\end{align}
where $\{\Step{\iter}\}_{\iter=0}^\infty$ is a sequence of positive
step sizes (to be chosen by the user).  We assume throughout that the
gradient descent procedure is initialized with $\BVEC{0} = 0$.

The parameter estimate $\BVEC{\iter}$ at iteration $t$ defines a
function estimate $\FUNIT{\iter}$ in the following way.  We first
compute\footnote{If the empirical matrix $\EmpKer$ is not invertible,
  then we use the pseudoinverse. Note that it may appear as though a
  matrix inversion is required to estimate $\origweight^{\iter}$ for
  each $\iter$ which is computationally intensive.  However, the
  weights $\origweight^\iter$ may be computed directly via the
  iteration $\origweight^{\iter+1} = \origweight^{\iter} -
  \Step{\iter} \EmpKer (\origweight^{\iter} -
  \frac{\Ydata}{\sqrt{\numobs}} )$.  However, the equivalent
  update~\eqref{EqnGradBvec} is more convenient for our analysis.} the
weight vector $\origweight^\iter = \sqrt{\EmpKer^{-1}} \;
\BVEC{\iter}$, which then defines the function estimate
$\FUNIT{\iter}(\cdot) = \frac{1}{\sqrt{\numobs}} \sum_{i=1}^\numobs
\origweight^\iter_i \Ker(\cdot, x_i)$ as before.  In this paper, our
goal is to study how the sequence $\{\FUNIT{\iter}\}_{\iter=0}^\infty$
evolves as an approximation to the true regression function $\fstar$.
We measure the error in two different ways: the $L^2(\mprob_\numobs)$
norm
\begin{align}
\label{EqnDefnEllTwoEmp}
\|f^\iter - \fstar\|_\numobs^2 & \defn \frac{1}{\numobs}
\sum_{i=1}^\numobs \big( f^\iter(x_i) - \fstar(x_i) \big)^2
\end{align}
compares the functions only at the observed design points, whereas the
$L^2(\mprob)$-norm
\begin{align}
\label{EqnDefnEllTwoPop}
\|f^\iter - \fstar\|_2^2 & \defn \Exs \Big[ \big(f^\iter(X) -
  \fstar(X) \big)^2 \Big]
\end{align}
corresponds to the usual mean-squared error.


\subsection{Overfitting and early stopping}

In order to illustrate the phenomenon of interest in this paper, we
performed some simulations on a simple problem.  In particular, we
formed $\numobs = 100$ i.i.d. observations of the form $y =
\fstar(x_i) + w_i$, where $w_i \sim N(0, 1)$, and using the fixed
design $x_i = i/\numobs$ for $i= 1, \ldots, \numobs$.  We then
implemented the gradient descent update~\eqref{EqnGradBvec} with
initialization $\BVEC{0} = 0$ and constant step sizes $\STEP{\iter} =
0l25$.  We performed this experiment with the regression function
$\fstar(x) = |x-1/2| - 1/2$, and two different choices of kernel
functions.  The kernel $\Ker(x, x') = \min \{x, x'\}$ on the unit
square $[0,1] \times [0,1]$ generates an RKHS of Lipschitz functions,
whereas the Gaussian kernel $\Ker(x, x') = \exp(-\frac{1}{2} (x -
x')^2)$ generates a smoother class of infinitely differentiable
functions.

\begin{figure}[h]
\begin{center}
\begin{tabular}{ccc}
\widgraph{.45\textwidth}{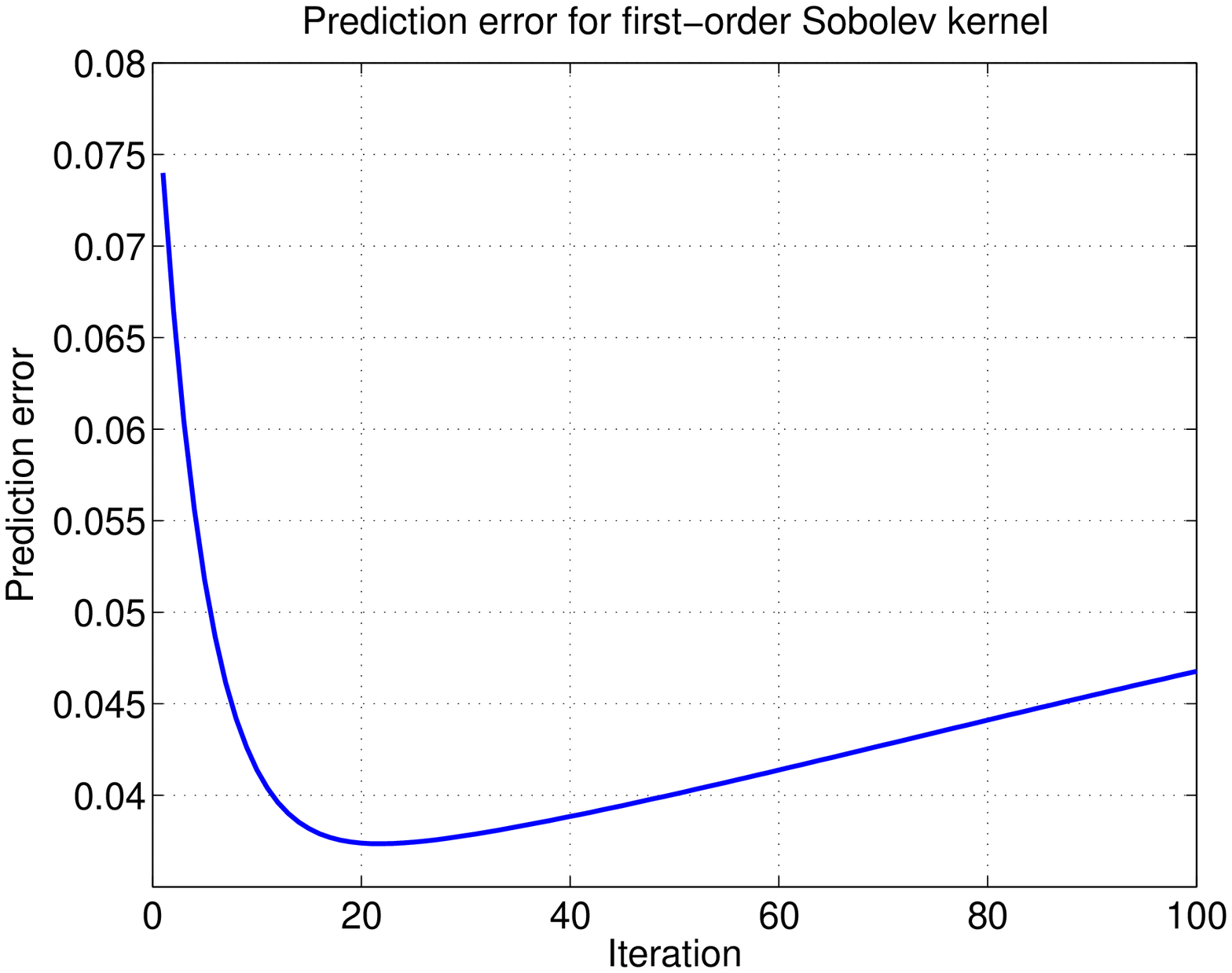} & & 
\widgraph{.45\textwidth}{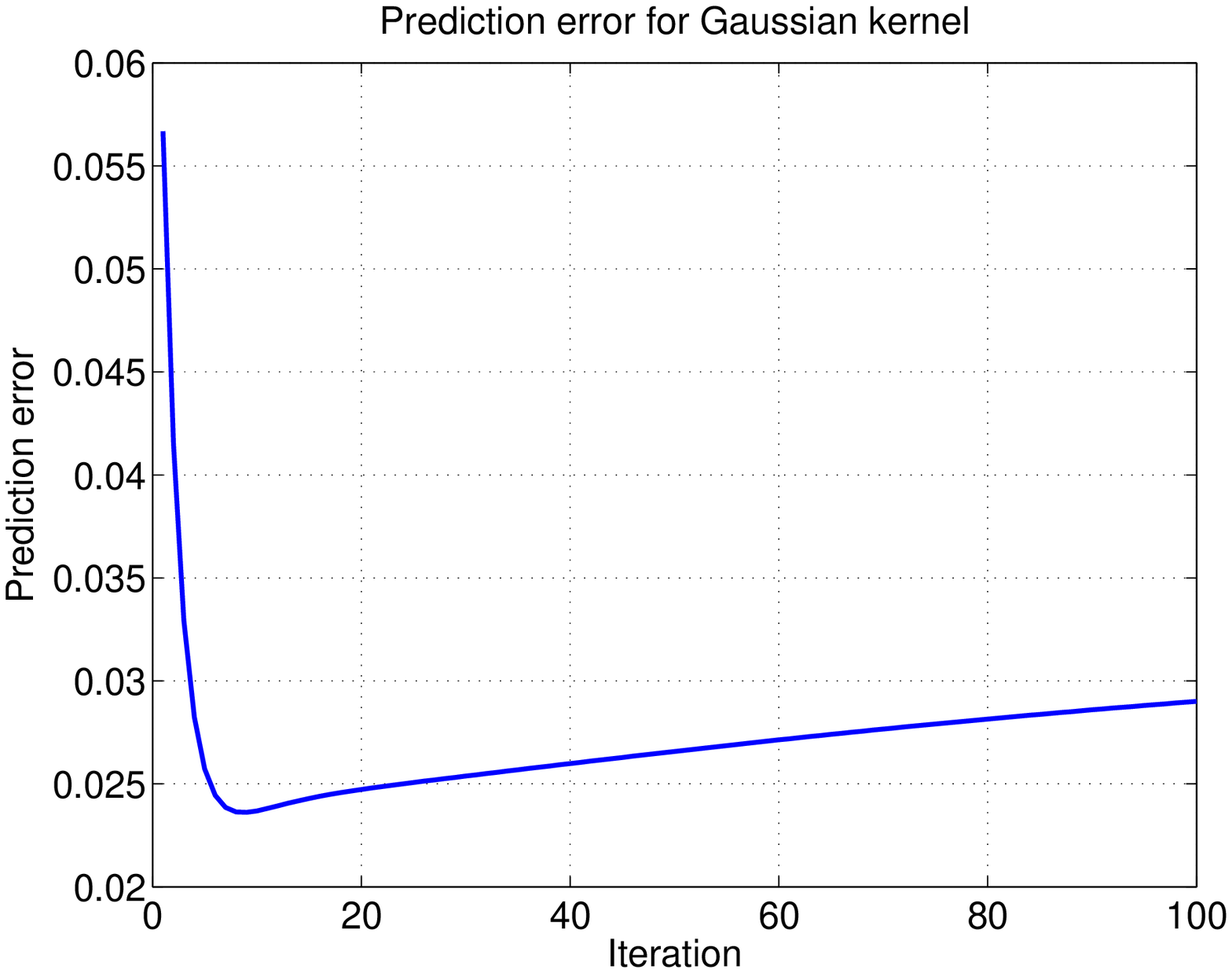} \\
(a) & & (b)
\end{tabular}
\caption{Behavior of gradient descent update~\eqref{EqnGradBvec} with
  constant step size \mbox{$\STEP{} = 0.25$} applied to least-squares
  loss with $\numobs = 100$ with equi-distant design points $x_i =
  i/\numobs$ for $i = 1, \ldots, \numobs$, and regression function
  $\fstar(x) = |x-1/2| - 1/2$.  Each panel gives plots the
  $L^2(\mprob_\numobs)$ error $\|\FUNIT{\iter} - \fstar\|_\numobs^2$
  as a function of the iteration number $\iter = 1, 2, \ldots, 100$.
  (a) For the first-order Sobolev kernel $\Ker(x, x') = \min\{x,
  x'\}$.  (b) For the Gaussian kernel $\Ker(x, x') = \exp(-\frac{1}{2}
  (x-x')^2)$. }
\label{FigPath}
\end{center}
\end{figure}

Figure~\ref{FigPath} provides plots of the squared prediction error
$\|\FUNIT{\iter} - \fstar\|_\numobs^2$ as a function of the iteration
number $\iter$.  For both kernels, the prediction error decreases
fairly rapidly, reaching a minimum before or around $T \approx 20$
iterations, before then beginning to increase.
As the analysis of this paper will clarify, too many iterations lead
to fitting the noise in the data (i.e., the additive perturbations
$w_i$), as opposed to the underlying function $\fstar$.  In a
nutshell, the goal of this paper is to quantify precisely the meaning
of ``too many'' iterations, and in a data-dependent and easily
computable manner.


\section{Main results and their consequences}
\label{SecMain}

In more detail, our main contribution is to formulate a data-dependent
stopping rule, meaning a mapping from the data $\{(x_i,
y_i)\}_{i=1}^\numobs$ to a positive integer $\STOP$, such that the two
forms of prediction error $\|\FUNIT{\STOP} - \fstar\|_\numobs$ and
$\|\FUNIT{\STOP} - \fstar\|_2$ are minimal.  In our formulation of
such a stopping rule, two quantities play an important role: first,
the \emph{running sum} of the step sizes
\begin{align}
\RUNSUM{t} & \defn \sum_{\tau=0}^{t-1}{\Step{\tau}},
\end{align}
and secondly, the eigenvalues $\kereigemp_1 \geq \kereigemp_2 \geq
\cdots \geq \kereigemp_\numobs \geq 0$ of the empirical kernel matrix
$\EmpKer$ previously defined~\eqref{EqnDefnEmpiricalKernel}.  The
kernel matrix and hence these eigenvalues are computable from the
data. We also note that there is a large body of work on fast
computation of kernel eigenvalues (e.g., see the
paper~\cite{DrinMah05} and references therein).

\subsection{Stopping rules and general error bounds}

Our stopping rule involves the use of a model complexity measure,
familiar from past work on uniform laws over kernel
classes~\cite{Bartlett02, Mendelson02}, known as the local empirical
Rademacher complexity.  For the kernel classes studied in this paper,
it takes the form
\begin{align}
\label{EqnDefnKernCompEmp}
\RadEmp_{\EmpKer}(\varepsilon) & \defn \biggr[ \frac{1}{n}
  \sum_{i=1}^n \min \big \{ \kereigemp_i, \varepsilon^2 \big \}
  \biggr]^{1/2}.
\end{align}
For a given noise variance $\sigma > 0$, a closely related
quantity---one of central importance to our analysis---is
\emph{critical empirical radius} $\EMPCRIT > 0$, defined to be the
smallest positive solution to the inequality
\begin{align}
\label{EqnDefnEmpCrit}
\RadEmp_{\EmpKer}(\varepsilon ) & \leq \varepsilon^2/(2 e \sigma).
\end{align}
The existence and uniqueness of $\EMPCRIT$ is guaranteed for any
reproducing kernel Hilbert space; see Appendix~\ref{AppRade} for
details.  As clarified in our proof, this inequality plays a key role
in trading off the bias and variance in a kernel regression estimate.

Our stopping rule is defined in terms of an analogous inequality that
involves the running sum \mbox{$\RUNSUM{\iter} = \sum_{\tau =
    0}^{t-1}{\Step{\tau}}$} of the step sizes. Throughout this paper,
we assume that the step sizes are chosen to satisfy the following
properties:
\begin{itemize}
\item Boundedness: $0 \; \leq \; \Step{\tau} \; \leq \; \min \{1,
  1/\kereigemp_1 \}$ for all $\tau = 0, 1, 2, \ldots$.
\item Non-increasing: $\Step{\tau+1} \leq \Step{\tau}$ for all $\tau =
  0, 1, 2, \ldots$.
\item Infinite travel: the running sum $\RUNSUM{\iter} =
  \sum_{\tau=0}^{\iter-1} \STEP{\tau}$ diverges as $\iter \rightarrow
  +\infty$.
\end{itemize}
We refer to any sequence $\{\STEP{\tau}\}_{\tau=0}^\infty$ that
satisfies these conditions as a \emph{valid stepsize sequence}.  We
then define the \emph{stopping time}
\begin{align}
\label{EqnStoppingRule}
\STOP & \defn \arg \min \biggr \{ \iter \in \Nat \, \mid
\RadEmp_{\EmpKer} \big(1/\sqrt{\RUNSUM{\iter}}\big) > (2 e \sigma
\RUNSUM{\iter})^{-1} \biggr \} - 1.
\end{align}
As discussed in Appendix~\ref{AppRade}, the integer $\STOP$ belongs to
the interval $[0, \infty)$ and is unique for any valid stepsize
  sequence.  As will be clarified in our proof, the intuition
  underlying the stopping rule~\eqref{EqnStoppingRule} is that the sum
  of the step-sizes $\RUNSUM{\iter}$ acts as a tuning parameter that
  controls the bias-variance tradeoff.  The stated choice of
  $\widehat{T}$ optimizes this trade-off.

The following result applies to any sequence $\{ \FUNIT{\iter}
\}_{\iter=0}^\infty$ of function estimates generated by the gradient
iteration~\eqref{EqnGradBvec} with a valid stepsize sequence. 

\btheos
\label{ThmMain}
Given the stopping time $\STOP$ defined by the
rule~\eqref{EqnStoppingRule}, there are universal positive constants
$(\gencon_1, \gencon_2)$ such that the following events both hold with
probability at least $1 -\gencon_1 \exp(-\gencon_2 \numobs
\EMPCRIT^2)$:
\begin{enumerate}
\item[(a)] For all iterations $\iter = 1, 2,...,\STOP$:
\begin{align}
\label{EqnGeneralBound}
 \| \FUNIT{\iter} - f^*\|_\numobs^2 \; \leq \;
 \frac{4}{e \, \RUNSUM{\iter} }.
\end{align}
\item[(b)] At the iteration $\STOP$ chosen according to the stopping
  rule~\eqref{EqnStoppingRule}, we have
\begin{align}
\label{BoundAtOpt}
\| \FUNIT{\STOP}- \fstar\|_\numobs^2 & \leq 12 \, \EMPCRIT^2.
\end{align}
\item[(c)] Moreover, for all $t > \STOP$,
\begin{align}
\label{BoundAfterOpt}
\mathbb{E}[\| \FUNIT{\iter}- \fstar\|_\numobs^2] & \geq
\frac{\sigma^2}{4} \RUNSUM{\iter}
\RadEmp_{\EmpKer}(\RUNSUM{\iter}^{-1/2}).
\end{align}
\end{enumerate}
\etheos 

\noindent \paragraph{Remarks:} Although the bounds (a) and (b) are
stated as high probability claims, a simple integration argument can
be used to show that the expected mean-squared error (over the noise
variables, with the design fixed) satisfies a bound of the form
\begin{align}
\label{EqnIterationsEbound}
\Exs \big[ \|\FUNIT{\iter} - \fstar\|_\numobs^2 \big] \leq \frac{4}{e
  \, \RUNSUM{\iter} }.
\end{align}
Moreover, as will be clarified in corollaries to follow,
Theorem~\ref{ThmMain} can be used to show that our stopping rule
provides minimax-optimal rates for various function classes.  The
interpretation of Theorem~\ref{ThmMain} is as follows: if the sum of
the step-sizes $\RUNSUM{\iter}$ remains below the threshold defined
by~\eqref{EqnStoppingRule}, applying the gradient
update~\eqref{EqnGradBvec} reduces the prediction error.  Moreover,
note that for Hilbert spaces with a larger kernel complexity, the
stopping time $\STOP$ is smaller, since fitting functions in a larger
class incurs a greater risk of overfitting.\\
 
In the case of random design $x_i \sim \mprob$, we can also provide
bounds on the $L^2(\mprob)$-error $\|\FUNIT{\iter} - \fstar\|_2$.  In
this setting, for the purposes of comparing to minimax lower bounds,
it is also useful to state some results in terms of the population
analog of the local empirical Rademacher
complexity~\eqref{EqnDefnKernCompEmp}, namely the quantity
\begin{align}
\label{EqnDefnPopKernComp}
\Rad_{\Ker}(\EPAR) & \defn \biggr[\frac{1}{n}\sum_{j = 1}^{\infty}
  \min \big \{ \kereig_j, \EPAR^2 \big \} \biggr]^{1/2}.
\end{align}
Using this complexity measure, we define the \emph{critical population
  rate} $\POPCRIT$ to be the smallest positive solution to the
inequality
\begin{align}
\label{EqnDefnPopCrit}
40 \: \Rad_{\Ker}(\EPAR) & \leq \frac{\EPAR^2}{\sigma}.
\end{align}
(Our choice of the pre-factor $40$ is for later theoretical
convenience.)  In contrast to the critical empirical rate $\EMPCRIT$,
this quantity is not data-dependent, since it depends on the
population eigenvalues of the RKHS $\Hil$.

\btheos[Random design]
\label{ThmRandDesign}
Suppose that the design variables $\{x_i\}_{i=1}^\numobs$ are sampled
i.i.d. according to $\mprob$. Then under the conditions of
Theorem~\ref{ThmMain}, there are universal positive constants
$\gencon_j, j = 1, 2, 3$ such that
\begin{align}
\|\FUNIT{\STOP} - f^*\|_2^2 \leq \gencon_3 \POPCRIT^2
\end{align}
with probability at least $1 - \gencon_1 \exp(-\gencon_2 \numobs
\EMPCRIT^2)$.
\etheos Theorems~\ref{ThmMain} and~\ref{ThmRandDesign} are general
results that apply to any reproducing kernel Hilbert space. Their
proofs involve combination of direct analysis of our iterative
update~\eqref{EqnGradBvec} combined with techniques from empirical
process theory and concentration of measure~\cite{vandeGeer,Ledoux01};
see Section~\ref{SecProofs} for the details.

To compare with the past work of B\"{u}hlmann and
Yu~\cite{BuhlmannYu03}, they also provide a theoretical analysis for
gradient descent (referred to as $L^2$-boosting in their paper),
focusing exclusively on the fixed design case.  Our theory applies to
random as well as fixed design, and a broader set of step-size
choices. The most significant difference between Theorem~\ref{ThmMain}
in our paper and Theorem 3 in the paper~\cite{BuhlmannYu03} is that we
provide a data-dependent stopping rule where as their analysis does
not lead to a computable stopping rule.


\subsection{Some consequences for specific kernel classes}

Let us now illustrate some consequences of our general theory for
special choices of kernels that are of interest in practice.

\paragraph{Kernels with polynomial eigendecay:}
We begin with the class of RKHSs whose eigenvalues satisfy a
\emph{polynomial decay condition}, meaning that
\begin{align}
\label{EqnPolyDecay}
\kereig_k & \leq C \big(\frac{1}{k}\big)^{2 \smooth} \qquad \mbox{for
  some $\smooth > 1/2$ and constant $C$.}
\end{align}
Among other examples, this type of scaling covers various types of
Sobolev spaces, consisting of functions with $\nu$ derivatives
(e.g.,~\cite{BirSol67,Gu02}).  For instance, the first-order Sobolev
kernel \mbox{$\Ker(x, x') = \min \{x, x'\}$} on the unit square
\mbox{$[0,1] \times [0,1]$} generates an RKHS of functions that are
differentiable almost \mbox{everywhere}, given by
\begin{align}
\label{EqnFirstOrderSobolev}
\Hil & \defn \big \{ f: [0,1] \rightarrow \real \mid \, f(0) = 0,
\quad \int_0^1 (f'(x))^2 dx < \infty \big \},
\end{align}
For the uniform measure on $[0,1]$, this class exhibits polynomial
eigendecay~\eqref{EqnPolyDecay} with $\smooth = 1$.  For any class
that satisfies the polynomial decay condition, we have the following
corollary: \\
\bcors
\label{CorAchieveSmooth}
Suppose that in addition to the assumptions of
Theorem~\ref{ThmRandDesign}, the kernel class $\Hil$ satisfies the
polynomial eigenvalue decay~\eqref{EqnPolyDecay} for some parameter
$\smooth > 1/2$.  Then there is a universal constant $\gencon_5$ such
that
\begin{align}
\label{EqnSobolevUpper}
\Exs \big[\|\FUNIT{\STOP}- f^*\|_2^2] & \leq \gencon_5 \,
\big(\frac{\sigma^2}{\numobs} \big)^{\frac{2 \smooth}{2 \smooth+1}}.
\end{align}
Moreover, if $\kereig_k \geq c \, (1/k)^{2 \smooth}$ for all $k = 1,
2, \ldots$, then
\begin{align}
\Exs \big[\|\FUNIT{\iter} - \fstar\|_2^2 \big] & \geq \frac{1}{4} \min
\big \{ 1, \; \sigma^2 \frac{(\RUNSUM{\iter})^{\frac{1}{2
      \smooth}}}{\numobs} \big \} \quad \mbox{for all iterations
  $\iter = 1, 2, \ldots$.}
\end{align}
\ecors

\noindent The proof, provided in
Section~\ref{SecProofCorAchieveSmooth}, involves showing that the
population critical rate~\eqref{EqnDefnPopKernComp} is of the order
$\order(\numobs^{- \frac{2 \smooth}{2 \smooth+1}})$.  By known results
on non-parametric regression~\cite{Sto85, YanBar99}, the error
bound~\eqref{EqnSobolevUpper} is minimax-optimal.

In the special case of the first-order spline
family~\eqref{EqnFirstOrderSobolev}, Corollary~\ref{CorAchieveSmooth}
guarantees that
\begin{align}
\label{EqnOptimalSobone}
\Exs[ \|\FUNIT{\STOP} - \fstar\|_2^2] & \precsim \big(
\frac{\sigma^2}{\numobs} \big)^{2/3}.
\end{align}
In order to test the accuracy of this prediction, we performed the
following set of simulations.  First, we generated samples from the
observation model 
\begin{align}
\label{EqnStandard}
y_i & = \fstar(x_i) + w_i, \qquad \mbox{for $i = 1, 2, \ldots,
  \numobs$},
\end{align}
where $x_i = i/\numobs$, and $w_i \sim N(0,\sigma^2)$ are i.i.d. noise
terms.  We present results for the function $\fstar(x) = |x - 1/2| -
1/2$, a piecewise linear function belonging to the first-order Sobelev
class.  For all our experiments, the noise variance $\sigma^2$ was set
to one, but so as to have a data-dependent method, this knowledge was
not provided to the estimator.  There is a large body of work on
estimating the noise variance $\sigma^2$ in non-parametric regression
(see e.g. Hall and Marron~\cite{HallMarron90}). For our simulations,
we use the simple estimator based on Hall and
Marron~\cite{HallMarron90}. They proved that their estimator is ratio
consistent, which is sufficient for our purposes.

\begin{figure}[h]
\begin{center}
\begin{tabular}{ccc}
\widgraph{.45\textwidth}{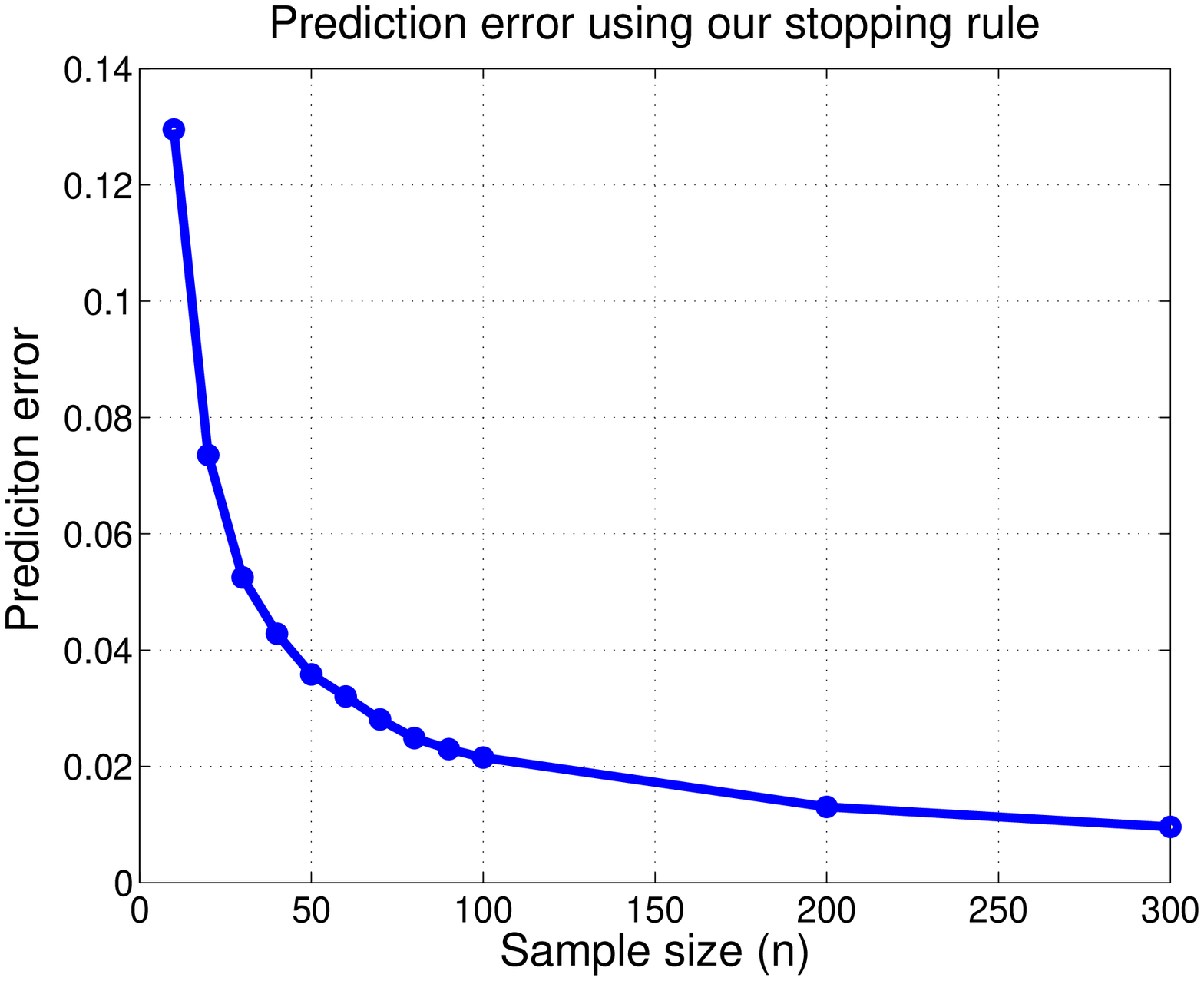} & & 
\widgraph{.45\textwidth}{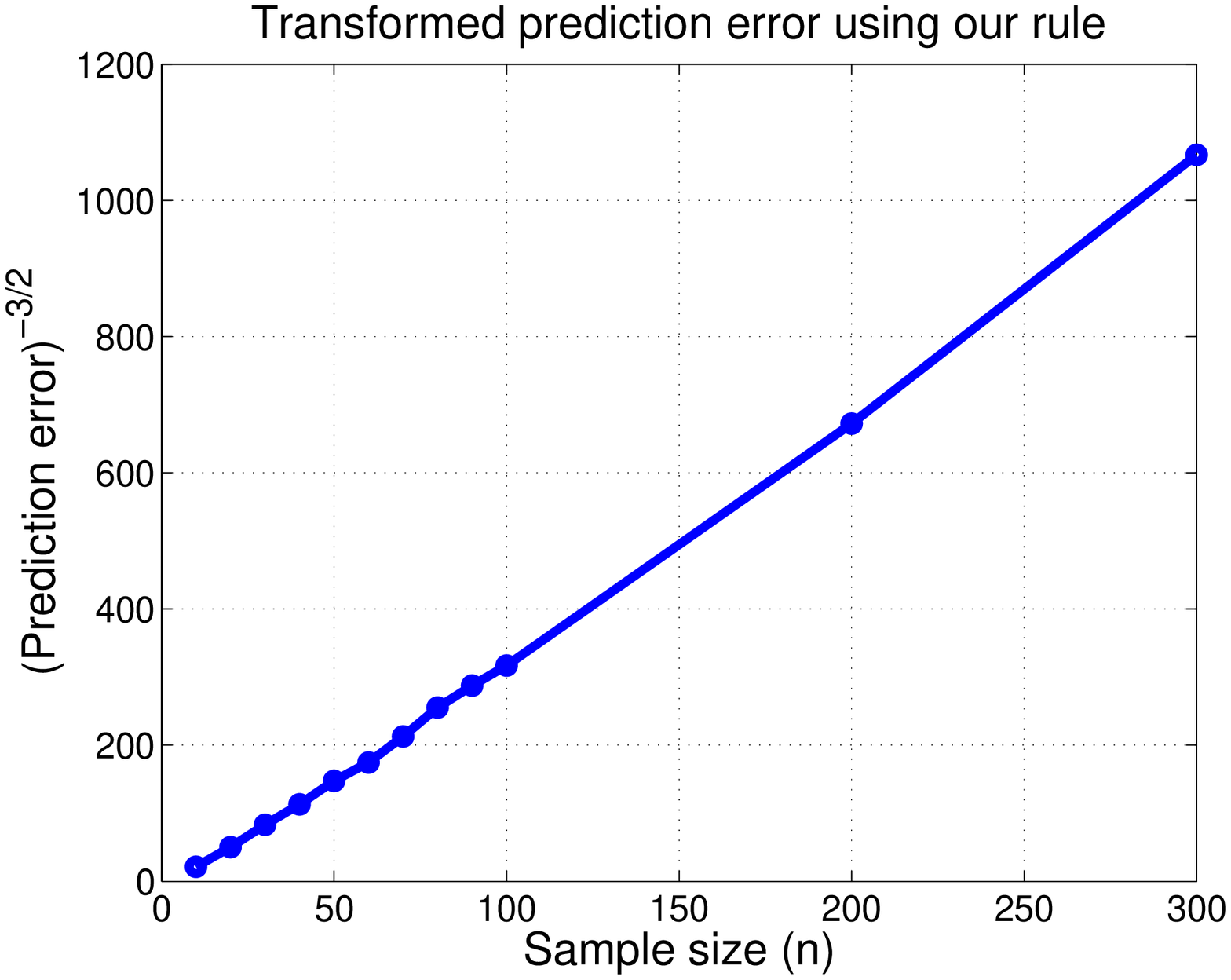} \\
(a) & & (b)
\end{tabular}
\caption{Prediction error obtained from the stopping rule applied to a
  regression with $\numobs$ samples of the form $\fstar(x_i) + w_i$ at
  equi-distant design points $x_i = i/\numobs$ for $i = 0, 1, \ldots
  99$, and i.i.d. Gaussian noise $w_i \sim N(0,)$.  For these
  simulations, the true regression function is given by $\fstar(x) =
  |x - 1/2| - 1/2$. Panel (a): Mean-squared error (MSE) using the
  stopping rule~\eqref{EqnStoppingRule} versus the sample size
  $\numobs$.  Each point is based on $10,000$ independent realizations
  of the noise variables $\{w_i\}_{i=1}^\numobs$.  $10,000$
  randomizations of $(w_i)_{i=1}^\numobs$ against the sample size
  $\numobs$. Panel (b): Plots of the quantity $MSE^{-3/2}$ versus
  sample size $\numobs$.  As predicted by the theory, this
  representation yields a straight line.}
\label{FigRates}
\end{center}
\end{figure}

 For a range of sample sizes $\numobs$ between $10$ and $300$, we
 performed the updates~\eqref{EqnGradBvec} with constant stepsize
 $\STEP{} = 0.25$, stopping at the specified time $\STOP$.  For each
 sample size, we performed $10,000$ independent trials, and averaged
 the resulting prediction errors.  In panel (a) of
 Figure~\ref{FigRates}, we plot the mean-squared error versus the
 sample size, which shows consistency of the method.  We also plotted
 the mean-squared error raised to the power $-3/2$ versus the sample
 size.  After this rescaling, the bound~\eqref{EqnOptimalSobone}
 predicts a linear relation, as is observed in panel (b) of
 Figure~\ref{FigRates}.  We also performed the same experiments for
 the case of randomly drawn designs $x_i \sim \mbox{Unif}(0,1)$.  In
 this case, we observed similar results but with more trials required
 to average out the additional randomness in the design.


\paragraph{Finite rank kernels:}  We 
now turn to the class of RKHSs based on finite-rank kernels, meaning
that there is some finite integer $\kerrank < \infty$ such that
$\kereig_j = 0$ for all $j \geq \kerrank + 1$.  For instance, the
kernel function $\Ker(x, x') = (1 + x x')^2$ is a finite rank kernel
(with $\kerrank = 2$) that generates the RKHS of all quadratic
functions.  More generally, for any integer $d \geq 2$, the kernel
$\Ker(x, x') = (1 + x x')^d$ generates the RKHS of all polynomials
with degree at most $d$.  For any such kernel, we have the following
corollary:

\bcors
\label{CorAchieveFinite}
If, in addition to the conditions of Theorem~\ref{ThmRandDesign}, the
kernel has finite rank $\kerrank$, then
\begin{align}
\Exs \big [\|\widehat{f}_{\widehat{T}}- f^*\|_2^2 \big] & \leq
\plaincon_5 \, \sigma^2 \frac{\kerrank}{\numobs}.
\end{align}

\ecors
\noindent Importantly, for a rank $\kerrank$-kernel, the rate
$\frac{\kerrank}{n}$ is minimax optimal in terms of squared $\LTP$
error~\cite[e.g.,]{RasWaiYu10b}. \\
%


\subsection{Comparison with other stopping rules}

\label{SecRuleComp}

In this section, we provide a comparison of our stopping rule to two
other stopping rules, as well as a oracle method (that involves
knowledge of $\fstar$, and so cannot be computed in practice).

\paragraph{Hold-out method:}

First, we consider a simple hold-out method: it performs gradient
descent using $50\%$ of the data, and uses the other $50 \%$ of the
data to estimate the risk (e.g.~\cite{DevroyeWagner79}).  Assuming
that the sample size is even for simplicity, we split the full data
set $\{x_i\}_{i=1}^\numobs$ into two equally sized subsets $\Strain$
and $S_{te}$.  The data indexed by the training set $\Strain$ is used
to estimate the function $\ftrain^t$ using the gradient descent
update~\eqref{EqnGradBvec}.  At each iteration $\iter = 0,1,2,
\ldots$, the data indexed by $S_{te}$ is used to estimate the risk via
$\RISKHO (f_\iter) = \frac{1}{n} \sum_{i \in S_{te}} \big(y_i -
\ftrain^\iter(x_i) \big)^2$, which defines the stopping rule
\begin{align}
\label{EqnStoppingRuleHO}
\STOPHO & \defn \arg \min \biggr \{ \iter \in \Nat \, \mid
\RISKHO(\ftrain^{\iter+1}) > \RISKHO(\ftrain^\iter) \biggr \} - 1.
\end{align}
A line of past work~\cite{Yao05,
  Bauer07,Caponnetto06,CaponnettoYao06,CaponnettoYaoJournal,
  DeVito10}) has analyzed stopping rules based on this type of
hold-out rule.  For instance, Caponetto~\cite{Caponnetto06} analyzes a
hold-out method, and shows that it yields rates that are optimal for
Sobolev spaces with $\smooth \leq 1$ but not in general. The major
drawback of using hold-out as that a percentage of the data is lost
which increases the risk.

\paragraph{SURE method:} Stein's Unbiased Risk estimate (SURE)
can be used to define an alternative stopping rule.  If we define the
shrinkage matrix $\tilde{S}^\iter = \prod_{\tau=0}^{\iter-1}{(I -
  \STEP{\tau}\EmpKer)}$, then it can be shown that the SURE
estimator~\cite{Stein81} takes the form
\begin{align}
\RISKSURE(f^\iter) & = \frac{1}{\numobs} \{ \numobs \sigma^2 + Y^T
(\tilde{S}^\iter)^2 Y - 2 \sigma^2 \trace(\tilde{S}^\iter) \},
\end{align}
which is easy to compute.  This risk estimate defines the associated
stopping rule
\begin{align}
\label{EqnStoppingRuleSURE}
\STOPSURE & \defn \arg \min \biggr \{ \iter \in \Nat \, \mid
\RISKSURE(f^{\iter+1}) > \RISKSURE(f^\iter) \biggr \} - 1.
\end{align}
In contrast with hold-out, this approach makes use of all the data.
However, we are not aware of any theoretical guarantees for early
stopping using the stopping rule~\eqref{EqnStoppingRuleSURE}.

It can be shown for both stopping rules~\eqref{EqnStoppingRuleHO} and
~\eqref{EqnStoppingRuleSURE}, a valid sequence of step-sizes
guarantees existence and uniqueness of the stopping point. Note that
our stopping rule $\STOP$ based on~\eqref{EqnStoppingRule} requires
estimation of both the empirical eigenvalues, and the noise variance
$\sigma^2$.  In contrast, the SURE-based rule requires estimation of
$\sigma^2$ but not the empirical eigenvalues, whereas the hold-out
rule requires no parameters to be estimated, but a percentage of the
data is used to estimate the risk.

\paragraph{Oracle method:}  As a third
point of reference, we also plot the mean-squared error for an
``oracle'' method.  It is allowed to base its stopping time on the
exact prediction error $\RISKOracle(f^\iter) = \|f^{\iter} -
\fstar\|_\numobs^2$, which defines the oracle stopping rule
\begin{align}
\label{EqnStoppingRuleOracle}
\STOPOracle & \defn \arg \min \biggr \{ \iter \in \Nat \, \mid
\RISKOracle(f^{\iter+1}) > \RISKOracle(f^\iter) \biggr \} - 1.
\end{align}
Note that this stopping rule is not computable from the data, since it
assumes exact knowledge of the function $\fstar$ that we are trying to
estimate.

In order to compare our stopping rule~\eqref{EqnStoppingRule} with
these alternatives, we generated i.i.d. samples from the previously
described model (see equation~\eqref{EqnStandard} and the following
discussion).  We varied the sample size $\numobs$ from $10$ to $300$,
and for each sample size, we performed $10,000$ independent trials
(randomizations of the noise variables $\{w_i\}_{i=1}^\numobs$), and
computed the average of squared prediction error.

\begin{figure}[h]
\begin{center}
\begin{tabular}{ccc}
\widgraph{.45\textwidth}{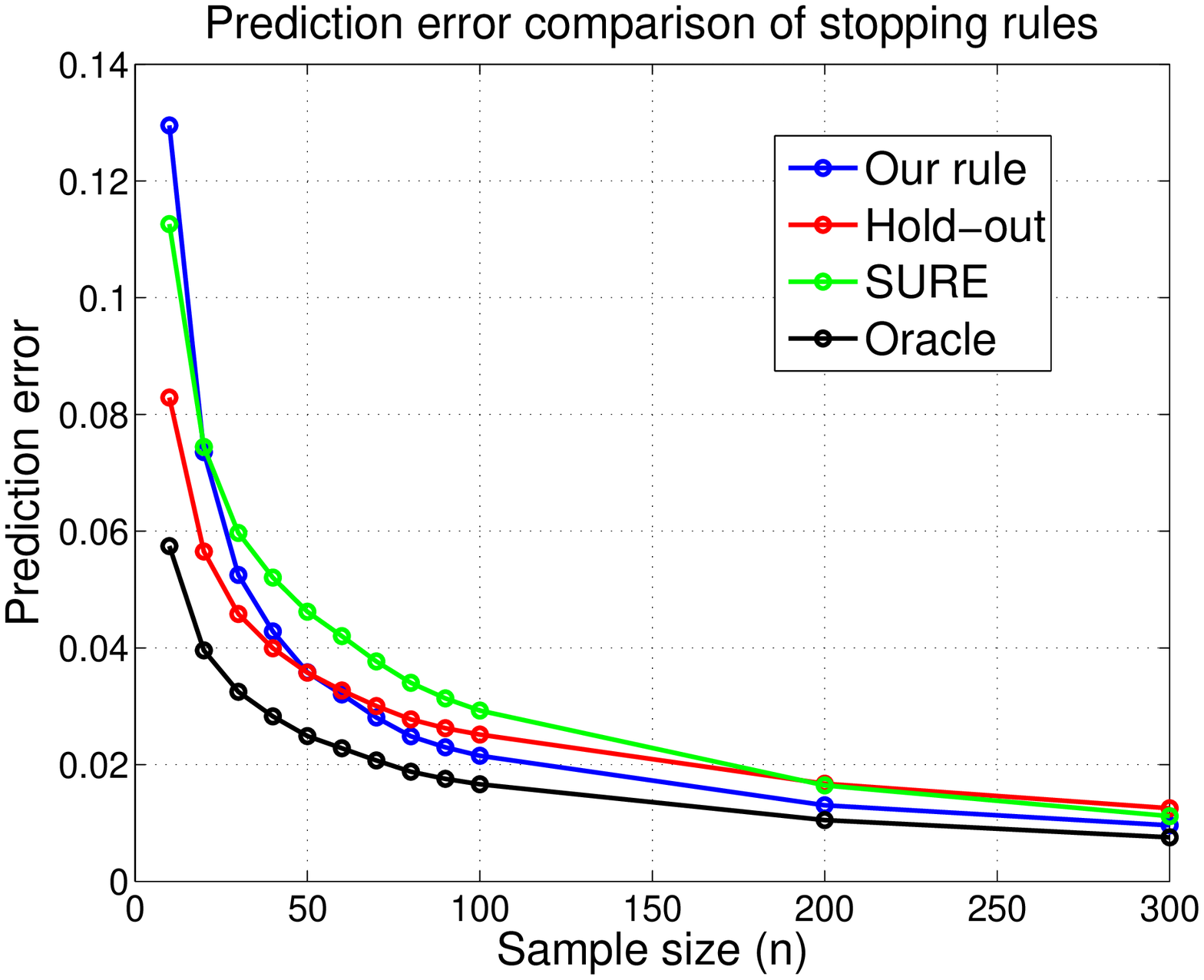} & &
\widgraph{.45\textwidth}{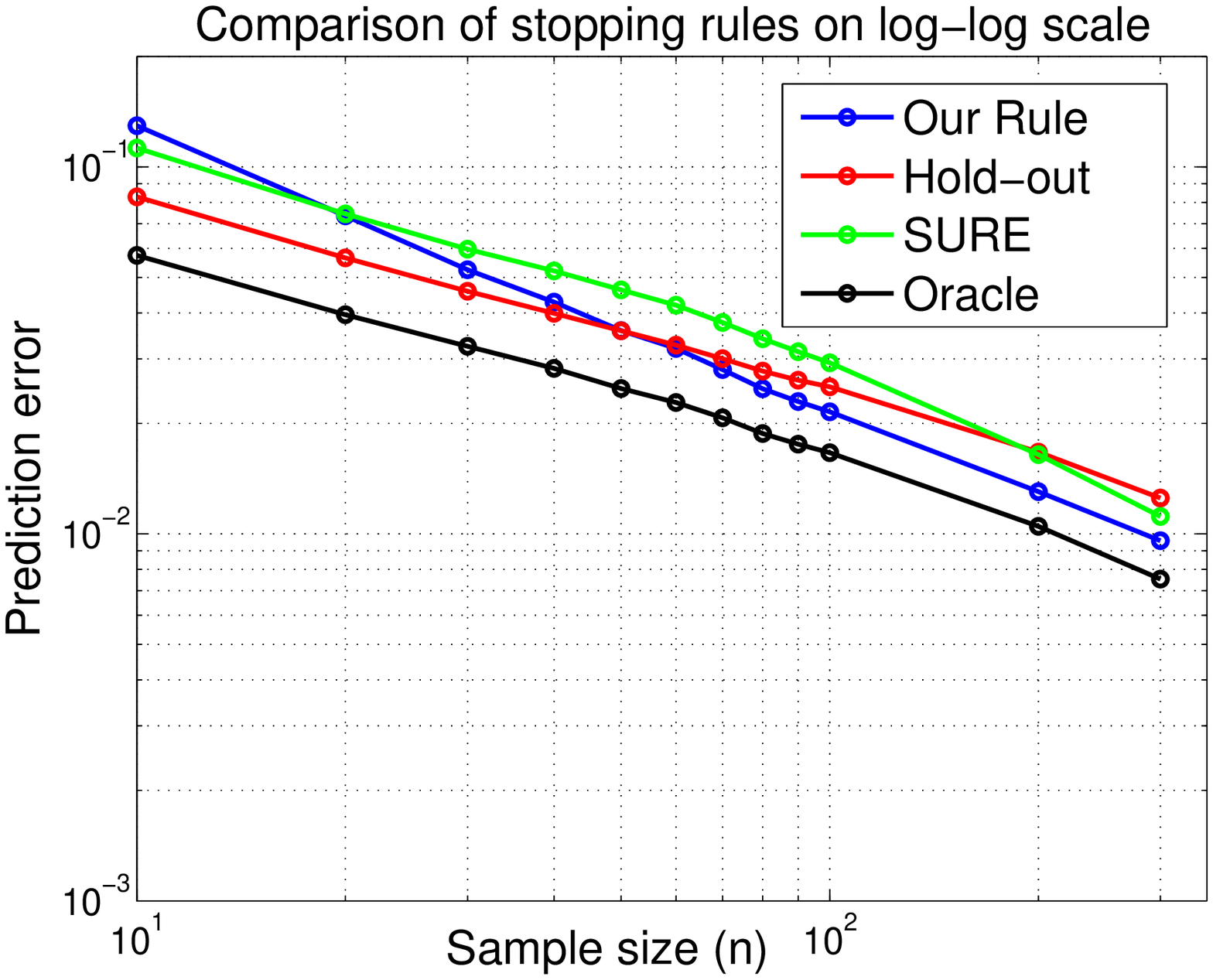} \\
(a) & & (b)
\end{tabular}
\caption{The non-parametric function is $f^*(x) = |x - 1/2| - 1/2$
  with kernel $\Ker(x,y) = \min(|x|,|y|)$. We apply the gradient
  descent update~\eqref{EqnGradBvec} with $\Step{\iter}=1$ for all
  $t$, and plot the average mean-squared error over $10,000$
  randomizations against the sample size for $n
  =10,20,30,40,50,60,70,80,90,100,200,300$. Mean-squared error is
  plotted for $4$ stopping rules: (i) our stopping
  rule~\eqref{EqnStoppingRule}; (ii) holding out $50 \%$ of the data
  and using~\eqref{EqnStoppingRuleHO}; (iii)
  SURE~\eqref{EqnStoppingRuleSURE}; and (iv) oracle stopping
  rule~\eqref{EqnStoppingRuleHO}. For panel (a) results are plotted on
  a normal scale and for panel (b), curves are plotted using a log-log
  scale.}
\label{FigHO}
\end{center}
\end{figure}
Figure~\ref{FigHO} plots the resulting mean-squared errors of our
stopping rule, the hold-out stopping rule~\eqref{EqnStoppingRuleHO},
the SURE-based stopping rule~\eqref{EqnStoppingRuleSURE}, and the
oracle rule~\eqref{EqnStoppingRuleOracle}.  Panel (a) shows the
mean-squared error versus sample size, whereas panel (b) shows the
same curves in terms of logarithm of mean-squared error.  Our proposed
rule exhibits better performance than the hold-out and SURE-based
rules for sample sizes $\numobs$ larger than $50$.  On the flip side,
since the construction of our stopping rule is based on the assumption
that $\fstar$ belongs to a known RKHS, it is unclear how robust it
would be to model mis-specification.  In contrast, the hold-out and
SURE-based stopping rules are generic methods, not based directly on
the RKHS structure, so might be more robust to model mis-specification.
Thus, one interesting direction is to explore the robustness of our
stopping rule.  On the theoretical front, it would be interesting to
determine whether the hold-out and/or SURE-based stopping rules can be
proven to achieve minimax optimal rates for general kernels, as we have
established for our stopping rule.


\subsection{Connections to  kernel ridge regression}
\label{SecRidgeCompare}

We conclude by presenting an interesting link between our early
stopping procedure and kernel ridge regression.  The kernel ridge
regression (KRR) estimate is defined as
\begin{align}
\label{EqnKernelRidge}	
\fhat_\PenPar & \defn \arg \min_{f \in \Hil} \big\{ \frac{1}{2
  \numobs} \sum_{i=1}^\numobs (y_i - f(x_i))^2 + \frac{1}{2 \PenPar}
\|f\|_\Hil^2 \big \},
\end{align}
where $\PenPar$ is the (inverse) regularization parameter.  For any
$\PenPar < \infty$, the objective is strongly convex, so that the KRR
solution is unique.

\begin{figure}[h]
\begin{center}
\begin{tabular}{ccc}
\widgraph{0.45\textwidth}{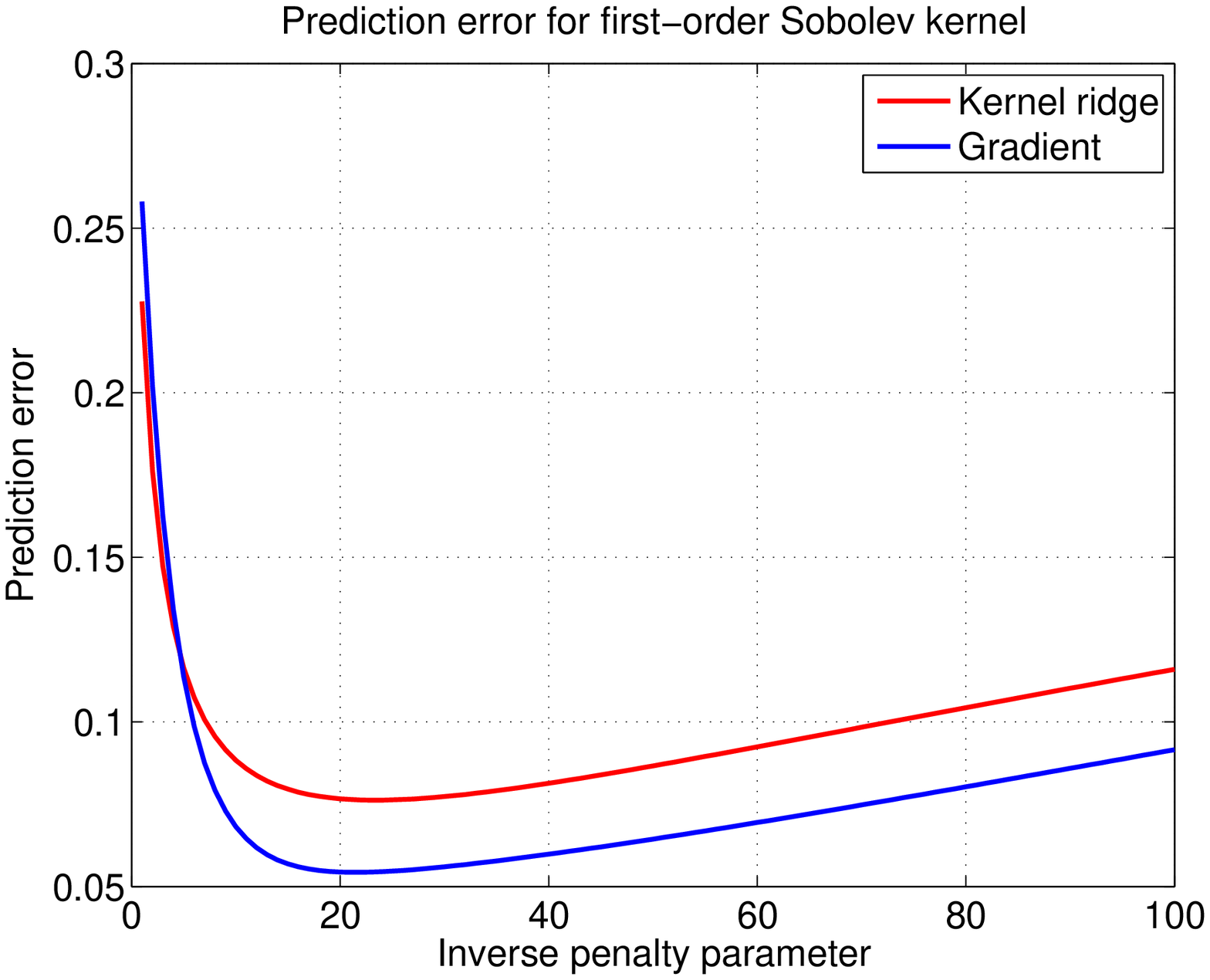} & & 
\widgraph{0.45\textwidth}{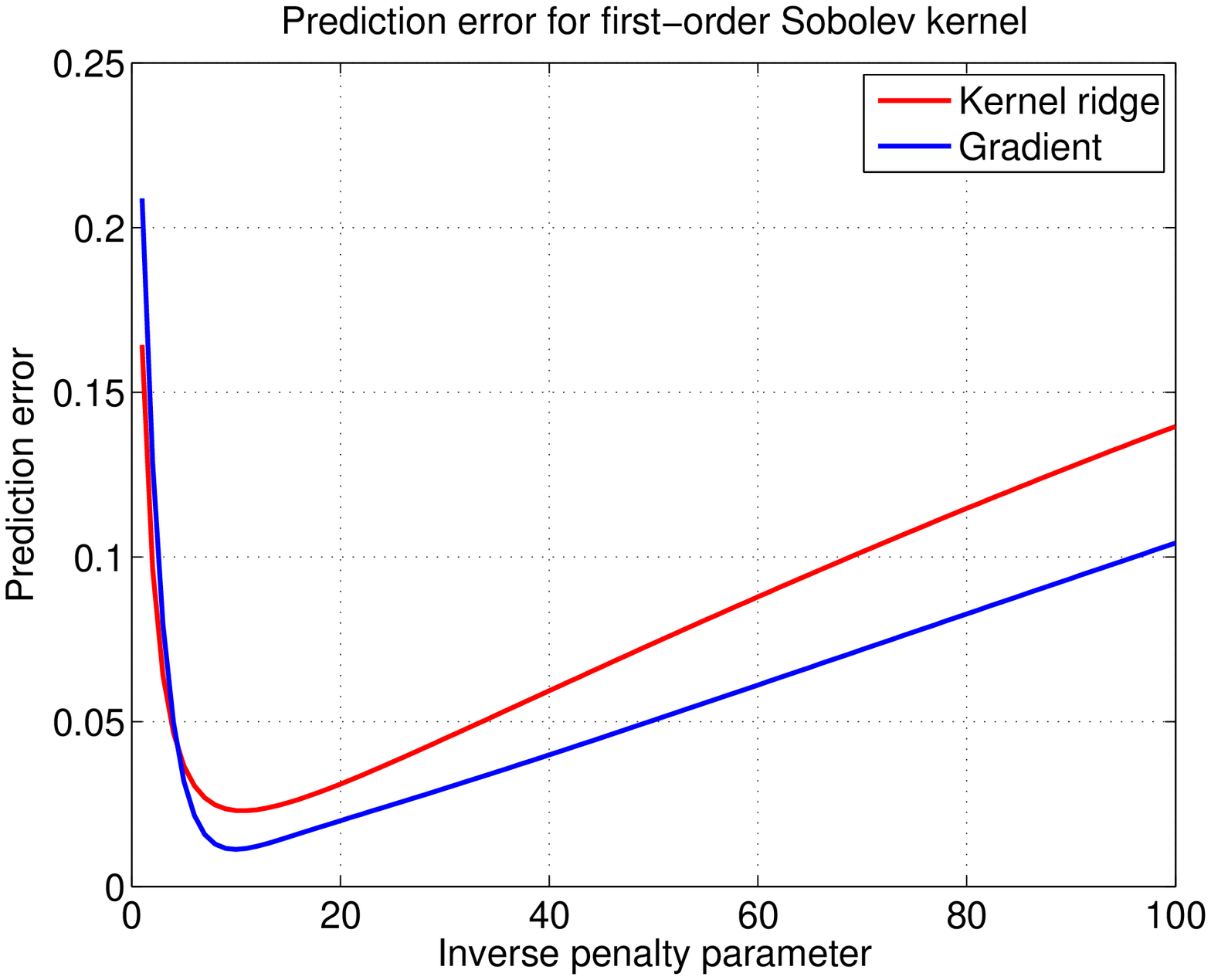} \\
(a) & & (b)
\end{tabular}
\caption{Comparison of the prediction error of the path of kernel
  ridge regression estimates~\eqref{EqnKernelRidge} obtained by
  varying $\PenPar \in [1, 100]$ to those of the gradient
  updates~\eqref{EqnGradBvec} over $100$ iterations with constant step
  size.  All simulations were performed with the kernel $\Ker(x, x') =
  \min \{x, x' \}$ based on $\numobs = 100$ samples at the design
  points $x_i = i/\numobs$ with $\fstar(x) = |x - 1/2| - 1/2$.  (a)
  Noise variance $\sigma^2 = 1$.  (b) Noise variance $\sigma^2 = 2$.}
\label{FigRidgePath}
\end{center}
\end{figure}

Friedman and Popescu~\cite{Friedman04} observed through simulations
that the regularization paths for early stopping of gradient descent
and ridge regression are similar, but did not provide any theoretical
explanation of this fact.  As an illustration of this empirical
phenomenon, Figure~\ref{FigRidgePath} compares the prediction error
$\|\fhat_{\PenPar} - \fstar\|_\numobs^2$ of the kernel ridge
regression estimate over the interval $\PenPar \in [1, 100]$ versus
that of the gradient update~\eqref{EqnGradBvec} over the first $100$
iterations.  Note that the curves, while not identical, are
qualitatively very similar.

From past theoretical work~\cite[e.g.,]{vandeGeer,Mendelson02}, kernel
ridge regression, with the appropriate setting of the penalty
parameter $\PenPar$, is known to achieve minimax-optimal error for
various kernel classes, among them the Sobolev and finite-rank kernels
for which stopping rule is provably optimal.  In this section, we
provide a theoretical basis for these connections, in particular by
showing that if the inverse penalty parameter $\PenPar$ is chosen
using the same criterion as our stopping rule, then the prediction
error satisfies the same type of bounds (with $\PenPar$ now playing
the role of the running sum $\RUNSUM{\iter}$).

More precisely, suppose that we choose $\PenParHat$ to be the smallest
positive solution to the inequality
\begin{align}
\label{EqnParChoice}
\big(4 \sigma \PenPar \big)^{-1} < \RadEmp_{\EmpKer}(1/\sqrt{\PenPar}
\big).
\end{align}
Note that this criterion is identical to the one underlying our
stopping rule, except that the continuous parameter $\PenPar$ replaces
the discrete parameter $\RUNSUM{\iter} =
\sum_{\tau=0}^{t-1}{\Step{\tau}}$.

\bprops
\label{PropMainRidge}
Consider the kernel ridge regression estimator~\eqref{EqnKernelRidge}
applied to $\numobs$ i.i.d. samples $\{(x_i, y_i)\}$ with $\sigma$-sub
Gaussian noise.  Then there are universal constants $(\gencon_1,
\gencon_2, \gencon_3)$ such that for all $\TAILPAR > 0$, the following
claims hold with probability at least $1 - \gencon_1 \exp(-\gencon_2
\, \numobs \, \EMPCRIT^2 )$:
\begin{enumerate}
\item[(a)] For all $0 < \PenPar \leq \PenParHat$, we have
\begin{align}
\label{EqnRidgeBoundBefore}
\|\fhat_{\PenPar}- f^*\|_\numobs^2 & \leq \frac{2}{\PenPar} 
\end{align}
\item[(b)] With $\PenParHat$ chosen according to the
  rule~\eqref{EqnParChoice}, we have
\begin{align}
\label{EqnRidgeBoundOpt}
\|\fhat_{\PenParHat} - \fstar\|_\numobs^2 & \leq \gencon_3 \:
\EMPCRIT^2.
\end{align}
\item[(c)] Moreover, for all $\PenPar > \PenParHat$, we have
\begin{align}
\label{EqnRidgeBoundOptAfter}
\mathbb{E}[\| \fhat_{\PenPar}- \fstar\|_\numobs^2] & \geq
\frac{\sigma^2}{4} \PenPar \RadEmp_{\EmpKer}(\PenPar^{-1/2}).
\end{align}
\end{enumerate}
\eprops

Note that (apart from a slightly different leading constant) the upper
bound~\eqref{EqnRidgeBoundBefore} is \emph{identical} to the upper
bound in equation~\eqref{EqnGeneralBound} in Theorem~\ref{ThmMain}.
The only difference is that the inverse regularization parameter
$\PenPar$ replaces the running sum $\RUNSUM{\iter} = \sum_{\tau =
  0}^{\iter-1}{\Step{\tau}}$.  Similarly, part (b) of
Proposition~\ref{PropMainRidge} guarantees that the kernel ridge
regression~\eqref{EqnKernelRidge} has prediction error that is upper
bounded by the empirical critical rate $\EMPCRIT^2$, as in part (b) of
Theorem~\ref{ThmMain}.  Let us emphasize that bounds of this type on
kernel ridge regression have been derived in past
work~\cite[e.g.,]{Mendelson02,vandeGeer}.  The novelty here is that
the structure of our result reveals the intimate connection to early
stopping, and in fact, the proofs follow a parallel thread.

In conjunction, Proposition~\ref{PropMainRidge} and
Theorem~\ref{ThmMain} provide a theoretical explanation for why, as
shown in Figure~\ref{FigRidgePath}, the paths of the gradient descent
update~\eqref{EqnGradBvec} and kernel ridge regression
estimate~\eqref{EqnKernelRidge} are so similar. However, it is
important to emphasize that from a computational point of view, early
stopping has certain advantages over kernel ridge regression.  In
general, solving a quadratic program of the
form~\eqref{EqnKernelRidge} requires on the order of
$\order(\numobs^3)$ basic operations, and this must be done repeatedly
at each new choice of $\PenPar$.  On the other hand, by its very
construction, the iterates of the gradient algorithm correspond to the
desired path of solutions, and each gradient update involves
multiplication by the kernel matrix, incurring $\order(\numobs^2)$
operations.


\section{Proofs}
\label{SecProofs}

We now turn to the proofs of our main results.  The main steps in each
proof are provided in the main text, with some of the more technical
results deferred to the appendix.

\subsection{Proof of Theorem~\ref{ThmMain}}

In order to derive upper bounds on the $\LTPn$-error in
Theorem~\ref{ThmMain}, we first rewrite the gradient
update~\eqref{EqnGradBvec} in an alternative form.  For each iteration
$\iter = 0, 1, 2, \ldots$, let us introduce the shorthand
\begin{align}
\FUN{\iter} & \defn \begin{bmatrix} f^\iter(x_1) & f^\iter(x_2) &
  \cdots & f^\iter(x_\numobs)
\end{bmatrix} \in \real^\numobs,
\end{align}
corresponding to the $\numobs$-vector obtained by evaluating the
function $f^\iter$ at all design points, and the short-hand
\begin{align}
\Wdata & \defn \begin{bmatrix} w_1, w_2,...,w_n\end{bmatrix} \in \real^\numobs,
\end{align}
corresponding to the vector of zero mean sub-Gaussian noise random variables. From
equation~\eqref{EqnRepresenter}, we have the relation 
\begin{align*}
\FUN{\iter} = \frac{1}{\sqrt{\numobs}} \EmpKer \, \origweight^\iter \;
= \; \frac{1}{\sqrt{\numobs}} \EmpKerRoot \, \BVEC{\iter}.
\end{align*}
Consequently, by multiplying both sides of the gradient
update~\eqref{EqnGradBvec} by $\EmpKerRoot$, we find that the sequence
$\{\FUN{\iter}\}_{\iter=0}^\infty$ evolves according to the recursion
\begin{align}
\label{EqnAltGrad}
\FUN{\iter+1} & = \FUN{\iter} - \Step{\iter} \EmpKer \, ( \FUN{\iter}
- \Ydata) \; = \; \Big( I_{\numobs \times \numobs} - \Step{\iter}
\EmpKer \Big) \FUN{\iter} - \Step{\iter} \EmpKer \, \Ydata.
\end{align}
Since $\BVEC{0} = 0$, the sequence is initialized with $\FUN{0} = 0$.
The recursion~\eqref{EqnAltGrad} lies at the heart of our analysis.

Letting $r = \mbox{rank}(\EmpKer)$, the empirical kernel matrix has
the eigendecomposition $\EmpKer = U \Dmat U^T$, where $U \in
\mathbb{R}^{\numobs \times \numobs}$ is an orthonormal matrix
(satisfying $U U^T = U^T U = I_{n \times n}$) and 
\begin{align*}
\Dmat \defn \diag(\kereigemp_1, \kereigemp_2, \ldots, \kereigemp_r,
0,0, \cdots, 0 )
\end{align*}
is the diagonal matrix of eigenvalues, augmented with $\numobs - r$
zero eigenvalues as needed.  We then define a sequence of diagonal
\emph{shrinkage matrices} $\SHRINK{\iter}$ as follows:
\begin{align*}
\SHRINK{\iter} & \defn \prod_{\tau = 0}^{t-1} {(I_{n \times n} -
  \Step{\tau} \Dmat)} \in \real^{n \times n}.
\end{align*}
The matrix $\SHRINK{\iter}$ indicates the extent of shrinkage towards
the origin; since $0 \; \leq \; \Step{\iter} \; \leq \; \min
\{1,1/\kereigemp_1 \}$ for all iterations $\iter$, in the positive
semodefinite ordering, we have the sandwich relation
\begin{align*}
0 \preceq \SHRINK{\iter + 1} \preceq \SHRINK{\iter} \preceq I_{\numobs
  \times \numobs}.
\end{align*}
Moreover, the following lemma shows that the
$L^2(\mprob_\numobs)$-error at each iteration can be bounded in terms
of the eigendecomposition and these shrinkage matrices:

\blems[Bias/variance decomposition]
\label{LemMain}
At each iteration $\iter = 0, 1, 2, \ldots$,
\begin{align}
\label{EqnMainUpper}
\| \FUNIT{\iter} - \fstar\|_\numobs^2
 & \leq \underbrace{\frac{2}{n}
  \sum_{j=1}^{r}{\SHRINKSQ{\iter}_{jj} [U^T \FUNSTAR]_j^2} +
  \frac{2}{n} \sum_{j=r+1}^{n}{[U^T \FUNSTAR]_j^2}}_{\mbox{Squared
    Bias $\BIASSQ$}} + \underbrace{\frac{2}{n} \sum_{j = 1}^{r} {(1 -
    \SHRINK{\iter}_{jj})^{2} [U^T \Wdata]_j^2}}_{\mbox{Variance
    $\MYVAR$}}.
\end{align}
Moreover, we have the lower bound $ \mathbb{E}[\| \FUNIT{\iter} -
\fstar\|_\numobs^2] \geq \mathbb{E}[\MYVAR]$.
\elems
\noindent See Appendix~\ref{AppLemMain} for the proof of this
intermediate claim. \\

In order to complete the proof of the upper bound in
Theorem~\ref{ThmMain}, our next step is to obtain high probability
upper bounds on these two terms.  We summarize our conclusions in an
additional lemma, and use it to complete the proof of
Theorem~\ref{ThmMain}(a) before returning to prove it.

\blems[Bounds on the bias and variance]
\label{LemBiasVarianceBound}
For all iterations $\iter = 1, 2, \ldots$, the squared bias is upper
bounded as
\begin{align}
\label{EqnBiasBound}
\BIASSQ & \leq \frac{1}{e \, \RUNSUM{\iter} },
\end{align}
Moreover, there is a universal constant $\gencon_1 > 0$ such that, for
any iteration $\iter = 1, 2, \ldots, \STOP$,
\begin{align}
\label{EqnVarBound}
\MYVAR \; \leq \; 5 \sigma^2 \, \RUNSUM{\iter} \Rad^2_{\EmpKer}
\big(1/\sqrt{\Par_\iter} \big)
\end{align}
with probability at least $1 - \exp \big(- \gencon_1 \, \numobs
\EMPCRIT^2 \big)$. Moreover for all $t$, we have $\mathbb{E}[\MYVAR]
\geq \frac{\sigma^2}{4} \, \RUNSUM{\iter} \Rad^2_{\EmpKer}\big(1/\sqrt{\Par_\iter} \big)$.
\elems

We can now complete the proof of Theorem~\ref{ThmMain}(a).  The
bound~\eqref{EqnGeneralBound} follows quickly: conditioned on the
event $\MYVAR \leq 5 \sigma^2 \RUNSUM{\iter} \Rad^2_{\EmpKer}
\big(1/\sqrt{\Par_\iter} \big)$, we have
\begin{align*} 
\|\FUNIT{\iter} - \fstar\|_\numobs^2 & \stackrel{(i)}{\leq} \BIASSQ +
\MYVAR \; \stackrel{(ii)}{\leq} \; \frac{1}{e \, \RUNSUM{\iter}} + 5
\sigma^2 \, \RUNSUM{\iter} \Rad^2_{\EmpKer}
\big(1/\sqrt{\RUNSUM{\iter}} \big) \stackrel{(iii)}{\leq} \frac{4}{e
  \, \RUNSUM{\iter}},
\end{align*}
where inequality (i) follows from~\eqref{EqnMainUpper} in
Lemma~\ref{LemMain}, and inequality (ii) follows from the bounds in
Lemma~\ref{LemBiasVarianceBound} and (iii) follows since $\iter \leq
\STOP$. The lower bound (c) in equation~\eqref{BoundAfterOpt} follows
from ~\eqref{EqnVarBound}.

Turning to the proof of part (b), using the upper bound from (a)
\begin{align*}
\|\FUNIT{\STOP} - \fstar\|_\numobs^2 & \leq \frac{1}{e \,
  \RUNSUM{\STOP}} + \frac{5}{\RUNSUM{\STOP}} \; \leq \;
\frac{4}{e \RUNSUM{\STOP}}.
\end{align*}
Based on the definition of $\STOP$ and $\EMPCRIT$,
we are guaranteed that $\frac{1}{\RUNSUM{\STOP+1}} \leq \EMPCRIT^2$,
Moreover, by the non-decreasing nature of our step sizes, we have
$\STEP{\STOP+1} \leq \STEP{\STOP}$, which implies that
$\RUNSUM{\STOP+1} \leq 2 \RUNSUM{\STOP}$, and hence
\begin{align*}
\frac{1}{\RUNSUM{\STOP}} \leq \frac{2}{\RUNSUM{\STOP+1}} \; \leq \; 2
\EMPCRIT^2.
\end{align*}
Putting together the pieces establishes the bound claimed in part (b).

It remains to establish the bias and variance bounds stated in
Lemma~\ref{LemBiasVarianceBound}, and we do so in the following
subsections.  The following auxiliary lemma plays a role in both
proofs:

\blems[Properties of shrinkage matrices]
\label{LemMatrices}
For all indices $j \in \{1, 2, \ldots, r\}$, the shrinkage matrices
$\SHRINK{\iter}$ satisfy the bounds
\begin{subequations}
\begin{align}
\label{EqnUpperBias}	
0 \leq \; \SHRINKSQ{\iter}_{jj} & \leq \frac{1}{2 e \RUNSUM{\iter}
  \kereigemp_j}, \quad \mbox{and} \\
\label{EqnLowerBias}		
\frac{1}{2} \min \{1, \Par_\iter \kereigemp_j \} & \leq
1-\SHRINK{\iter}_{jj} \; \leq \min \{ 1, \RUNSUM{\iter} \kereigemp_j
\}.
\end{align}
\end{subequations}
\elems
\noindent See Appendix~\ref{AppLemMatrices}  for the proof of this result.


\subsubsection{Bounding the squared bias}
\label{SecBias}

Let us now prove the upper bound~\eqref{EqnBiasBound} on the squared
bias.  We bound each of the two terms in the
definition~\eqref{EqnMainUpper} of $\BIASSQ$ in term.  Applying the
upper bound ~\eqref{EqnUpperBias} from Lemma~\ref{LemMatrices}, we see
that
\begin{align*}
\frac{2}{n} \sum_{j=1}^{r} {\SHRINKSQ{\iter}_{jj} [U^T
    \fstar(x_1^\numobs)]_j^2} & \leq \frac{1}{e \; \numobs \;
  \RUNSUM{\iter}} \; \sum_{j=1}^{r} { \frac{[U^T \fstar(x_1^\numobs)
    ]_j^2}{\kereigemp_j}}.
\end{align*}
Now consider the linear operator \mbox{$\Phi_X: \ell^2(\Nat)
  \rightarrow \real^\numobs$} defined element-wise via $[\Phi_X]_{jk}
= \phi_j(x_k)$.  Similarly, we define a (diagonal) linear operator
\mbox{$\DiagOpt: \ell^2(\Nat) \rightarrow \ell^2(\Nat)$} with entries
$[\DiagOpt]_{jj} = \lambda_j$ and $[\DiagOpt]_{jk} = 0$ for $j \neq
k$.  With these definitions, the vector $\FUN{} \in \real^\numobs$ can
be expressed in terms of some sequence $a \in \ell^2(\Nat)$ in the
form
\begin{align*}
\FUN{} & = \Phi_X \DiagOpt^{1/2} a.
\end{align*}
In terms of these quantities, we can write $\EmpKer =
\frac{1}{n}\Phi_X \DiagOpt \Phi_X^T$.  Moreover, as previously noted,
we also have $\EmpKer = U \Dmat U^T$ where $\Dmat = \diag
\{\kereigemp_1, \kereigemp_2, \ldots, \kereigemp_\numobs \}$, and $U
\in \real^{\numobs \times \numobs}$ is orthonormal.  Combining the two
representations, we conclude that
\begin{equation*}
\frac{\Phi_X \DiagOpt^{1/2}}{\sqrt{n}} = U \Dmat^{1/2} \Psi^*,
\end{equation*}	
for some linear operator $\Psi: \real^n \rightarrow \ell^2(\Nat)$
(with adjoint $\Psi^*$) such that $\Psi^* \Psi = I_{\numobs \times
  \numobs}$.  Using this equality, we have
\begin{align}
\frac{1}{e \, \RUNSUM{\iter} \, n} \sum_{j=1}^{r}{\frac{[U^T
      f^*(X)]_j^2}{\widehat{\Eig_j}}} & = \frac{1}{e \, \RUNSUM{\iter}
  \, \numobs} \; \sum_{j=1}^{r} {\frac{[U^T \Phi_X \DiagOpt^{1/2}
      a]_j^2}{\kereigemp_j}} \nonumber \\
& = \frac{1}{e \, \RUNSUM{\iter}} \; \sum_{j = 1}^{r} { \frac{[U^T U
           \Dmat^{1/2} V^* a ]_j^2}{\kereigemp_j}} \nonumber \\
& = \frac{1}{e \, \RUNSUM{\iter}} \; \sum_{j = 1}^{r}
     {\frac{\kereigemp_j \, [\Psi^* a]_j^2}{\kereigemp_j}} \nonumber \\
& \leq \frac{1}{e \, \RUNSUM{\iter}} \; \|\Psi^* a\|_2^2 \nonumber \\
\label{EqnBoundOne}
& \leq \frac{1}{e \, \RUNSUM{\iter}},
\end{align}
Here the final step follows from the fact that $\Psi$ is a unitary
operator, so that \mbox{$\|\Psi^* a\|_2^2 \leq \|a\|_2^2 =
  \|f^*\|_{\Hil}^2 \leq 1$.}

Turning to the second term in the definition~\eqref{EqnMainUpper}, we
have
\begin{align}
\sum_{j=r+1}^{n} {[U^T \fstar(x_1^\numobs) ]_j^2} & = \frac{2}{n}
\sum_{j=r+1}^{n}{[U^T \Phi_X \DiagOpt^{1/2} a]_j^2} \nonumber \\
& = \sum_{j=r+1}^\numobs {[U^T U \Dmat^{1/2} \Psi^* a]_j^2} \nonumber \\
& = \sum_{j = r+1}^{n} [\Dmat^{1/2} \Psi^* a]_j^2 \nonumber \\
\label{EqnBoundTwo}
& = 0,
\end{align}
where the final step uses the fact that $\Dmat^{1/2}_{jj} = 0$ for all
$j \in \{r +1, \ldots, \numobs \}$ by construction.  Combining the
upper bounds~\eqref{EqnBoundOne} and~\eqref{EqnBoundTwo} with the
definition~\eqref{EqnMainUpper} of $\BIASSQ$ yields the
claim~\eqref{EqnBiasBound}.


\subsubsection{Controlling the variance}
\label{SecVar}

Let us now prove the bounds~\eqref{EqnVarBound} on the variance term
$\MYVAR$.  (To simplify the proof, we assume throughout that $\sigma =
1$; the general case can be recovered by a simple rescaling argument).
By the definition of $\MYVAR$, we have
\begin{align*}
\MYVAR & = \frac{2}{n} \sum_{j = 1}^{r}{(1-\SHRINK{\iter}_{jj})^{2}
  [U^T \noise]_j^2} \; = \; \frac{2}{\numobs} \trace(U Q U^T \, w w^T),
\end{align*}
where $Q = \diag \{ (1 -\SHRINK{\iter}_{jj})^2, \; j = 1, \ldots,
\numobs \}$ is a diagonal matrix.  Since $\Exs[w w^T] \leq I_{\numobs
  \times \numobs}$ by assumption, we have $\Exs[\MYVAR] =
\frac{2}{\numobs} \trace(Q)$.  Using the upper bound in
equation~\eqref{EqnLowerBias} from Lemma~\ref{LemMatrices}, we have
\begin{align*}
\frac{1}{\numobs} \trace(Q) & \leq \frac{1}{\numobs} \sum_{j=1}^r \min
\{1, (\RUNSUM{\iter} \kereigemp_j)^2 \} \; = \; \RUNSUM{\iter} \;
\biggr( \Rad_{\EmpKer}(1/\sqrt{\RUNSUM{\iter}}) \biggr)^2,
\end{align*}
where the final equality uses the definition of $\Rad_\EmpKer$.
Putting together the pieces, we see that
\begin{subequations}
\begin{align}
\label{EqnUpperEmyvar}
\Exs[\MYVAR] & \leq 2 \; \RUNSUM{\iter} \biggr(
\Rad_{\EmpKer}(1/\sqrt{\RUNSUM{\iter}}) \biggr)^2.
\end{align}
Similarly, using the lower bound in equation~\eqref{EqnLowerBias}, we
can show that
\begin{align}
\label{EqnLowerEmyvar}
\Exs[\MYVAR] & \geq \frac{\sigma^2}{4} \RUNSUM{\iter} \biggr(
\Rad_{\EmpKer}(1/\sqrt{\RUNSUM{\iter}}) \biggr)^2.
\end{align}
\end{subequations}

Our next step is to obtain a bound on the two-sided tail probability
$\mprob[|\MYVAR - \Exs[\MYVAR]| \geq \delta]$, for which we make use of
a result on two-sided deviations for quadratic forms in sub-Gaussian
variables.  In particular, consider a random variable of the form
$Q_\numobs = \sum_{i,j=1}^\numobs a_{ij} (Z_i Z_j - \Exs[Z_i Z_j])$
where $\{Z_i\}_{i=1}^\numobs$ are i.i.d. zero-mean and sub-Gaussian
variables (with parameter $1$).  Wright~\cite{Wri73} proves that there
is a constant $c$ such that 
\begin{align}
\label{EqnWrightBound}
\mprob \big[|Q - \Exs[Q]| \geq \TAILPAR \big] \leq \exp \Big(-c \;
\min \big \{ \frac{\TAILPAR}{\opnorm{A}},
\frac{\TAILPAR^2}{\fronorm{A}^2} \big \} \Big) \quad \mbox{for all $u>
  0$,}
\end{align}
where $(\opnorm{A}, \fronorm{A})$ are (respectively) the operator and
Frobenius norms of the matrix \mbox{$A = \{a_{ij}\}_{i,j=1}^\numobs$.}

If we apply this result with $A= \frac{2}{\numobs} U Q U^T$ and $Z_i =
w_i$, then we have $Q = \MYVAR$, and moreover
\begin{align*}
\opnorm{A} & \leq \frac{2 }{\numobs}, \quad \mbox{and} \\
\fronorm{A}^2 \; = \; \frac{4}{\numobs^2} \trace(U^T Q U^T U Q U^T) &
= \frac{4}{\numobs^2} \trace(Q^2) \leq \frac{4}{\numobs^2} \trace(Q)
\leq \frac{4}{\numobs} \RUNSUM{\iter} \; \biggr(
\Rad_{\EmpKer}(1/\sqrt{\RUNSUM{\iter}}) \biggr).
\end{align*}
Consequently, the bound~\eqref{EqnWrightBound} implies that
\begin{align}
\label{EqnTwoMyvar}
\mprob \big[ |\MYVAR - \Exs[\MYVAR] | \geq \TAILPAR \big] & \leq \exp
\big( - 4 c \, \numobs \, \delta \min \{1, \delta \biggr(\RUNSUM{\iter} \Rad_{\EmpKer}(1/\sqrt{\RUNSUM{\iter}}) \biggr)^{-1}\} \big).
\end{align}
Since $\iter \leq \STOP$ setting $\delta = 3 \sigma^2 \RUNSUM{\iter} \; \biggr(\Rad_{\EmpKer}(1/\sqrt{\RUNSUM{\iter}}) \biggr)$, the claim~\eqref{EqnVarBound} follows.


\subsection{Proof of Theorem~\ref{ThmRandDesign}}

This proof is based on the following two steps:
\begin{itemize}
\item first, proving that the error $\|\FUNIT{\STOP} - \fstar\|_2$ in
  the $L^2(\mprob)$ norm is, with high probability, close to the error
  in the $L^2(\mprob_\numobs)$ norm, and
\item second, showing the empirical critical radius $\EMPCRIT$ defined
  in equation~\eqref{EqnDefnEmpCrit} is upper bounded by the
  population critical radius $\POPCRIT$ defined in
  equation~\eqref{EqnDefnPopCrit}.
\end{itemize}

\noindent Our proof is based on a number of more technical auxiliary
lemmas, proved in the appendices.  The first lemma provides a high
probability bound on the Hilbert norm of the estimate $\FUNIT{\STOP}$.
\blems
\label{LemHilbBound}
There exist universal constants $c_1$ and $c_2 > 0$ such that
$\|\FUNIT{\iter}\|_{\Hil} \leq 2$ for all $\iter \leq \STOP$ with
probability greater than or equal to $1 - c_1 \exp(-c_2 \numobs
\EMPCRIT^2)$.
\elems 
See Appendix~\ref{AppLemHilbBound} for the proof of this claim.  Our
second lemma shows in any bounded RKHS, the $\LTP$ and $\LTPn$ norms
are uniformly close up to the population critical radius $\POPCRIT$
over a Hilbert ball of constant radius:
\blems

\label{LemGeneral}
Consider a Hilbert space such that $\|g\|_\infty \leq \fbou$ for all
$g \in \Ball_\Hil(3)$.  Then there exist universal constants
$(\plaincon_1, \plaincon_2, \plaincon_3)$ such that for any $t \geq
\POPCRIT$, we have
\begin{equation}
\label{EqnSandwich}
| \|g\|_\numobs^2 - \|g\|_2^2 | \; \leq \; \plaincon_1 t^2,
\end{equation}
with probability at least $1 - \plaincon_2 \exp(-\plaincon_3 \numobs
t^2)$.
\elems
\noindent This claim follows from known results on reproducing kernel
Hilbert spaces (e.g., Lemma 5.16 in the paper~\cite{vandeGeer} and
Theorem 2.1 in the paper~\cite{Bartlett05}).  Our final lemma, proved
in Appendix~\ref{AppLemEmpTru}, relates the critical empirical radius
$\EMPCRIT$ to the population radius $\POPCRIT$:
\blems
\label{LemEmpTru}
The inequality $\EMPCRIT \leq \POPCRIT$ holds with probability at
least $1 - \plaincon_1 \exp( - \plaincon_2 \numobs \EMPCRIT^2)$.
\elems

With these lemmas in hand, the proof of the theorem is
straightforward.  First, from Lemma~\ref{LemHilbBound}, we have
$\|\FUNIT{\STOP}\|_{\Hil} \leq 2$ and hence by triangle inequality,
$\|\FUNIT{\STOP} - \fstar\|_{\Hil} \leq 3$ with high probability as
well.  Next, applying Lemma~\ref{LemGeneral} with $t = \POPCRIT$, we
find that
\begin{align*}
\| \FUNIT{\STOP} - \fstar \|_2^2 & \leq \| \FUNIT{\STOP} - \fstar
\|_\numobs^2 + \plaincon_1 \POPCRIT^2 \leq \plaincon_4(\EMPCRIT^2 +
\POPCRIT^2),
\end{align*}
with probability greater than $1 -\plaincon_2 \exp(-\plaincon_3
\numobs \POPCRIT^2)$.  Finally, applying Lemma~\ref{LemEmpTru} yields
that the bound \mbox{$\|\FUNIT{\STOP} - \fstar\|_2^2 \leq \plaincon
  \POPCRIT^2$} holds with the claimed probability.


\subsection{Proof of Corollaries}
\label{SecProofCorAchieveSmooth}

In each case, it suffices to upper bound the population critical rate
$\POPCRIT^2$ previously defined.

\paragraph{Proof of Corollary~\ref{CorAchieveFinite}:}

In this case, we have
\begin{align*}
\Rad_{\Ker}(\epsilon) & = \frac{1}{\sqrt{\numobs}} \sqrt{\sum_{j=1}^m
  \min \{ \kereig_j, \epsilon^2 \}} \; \leq \;
\sqrt{\frac{m}{\numobs}} \, \epsilon
\end{align*}
so that $\POPCRIT^2 = c' \sigma^2 \frac{m}{\numobs}$.

\paragraph{Proof of Corollary~\ref{CorAchieveSmooth}:}

For any $M \geq 1$, we have
\begin{align*}
\Rad_{\Ker}(\epsilon) \; = \; \frac{1}{\sqrt{\numobs}} \,
\sqrt{\sum_{j=1}^\infty \min \{C \, j^{-2 \smooth}, \epsilon^2 \}} &
\leq \sqrt{\frac{M}{\numobs}} \epsilon + \sqrt{\frac{C}{\numobs}} \,
\sqrt{\sum_{j=\lceil M \rceil}^\infty j^{-2 \smooth}} \\ & \leq
\sqrt{\frac{M}{\numobs}} \epsilon + \sqrt{\frac{C'}{\numobs}}
\sqrt{\int_{M}^\infty t^{-2 \smooth} dt} \\
& \leq \sqrt{\frac{M}{\numobs}} \epsilon + C''
\frac{1}{\sqrt{\numobs}} (1/M)^{\smooth -\frac{1}{2}}
\end{align*}
Setting $M = \epsilon^{-1/\smooth}$ yields $\Rad_\Ker(\epsilon) \leq
C^* \epsilon^{1 - \frac{1}{2 \smooth}}$.  Consequently, the critical
inequality $\Rad_\Ker(\epsilon) \leq 40 \epsilon^2/\sigma$ is
satisfied for $\POPCRIT \asymp (\sigma^2/\numobs)^{\frac{2 \smooth}{2
    \smooth+1}}$, as claimed.


\subsection{Proof of Proposition~\ref{PropMainRidge}}

We now turn to the proof of our results on the kernel ridge regression
estimate~\eqref{EqnKernelRidge}.  The proof follows a very similar
structure to that of Theorem~\ref{ThmMain}.  Recall the
eigendecomposition $\EmpKer = U \Dmat U^T$ of the empirical kernel
matrix, and that we use $r$ to denote its rank.  For each $\PenPar >
0$, we define the \emph{ridge shrinkage matrix}
\begin{align}
\label{EqnDefnKERSHRINK}
\KERSHRINK & \defn \big(I_{\numobs \times \numobs} + \PenPar \Dmat
\big)^{-1}.
\end{align}
We then have the following analog of Lemma~\ref{LemBiasVarianceBound}
from the proof of Theorem~\ref{ThmMain}:
\blems[Bias/variance decomposition for kernel ridge regression] 
\label{LemRidgeOne}
For any $\PenPar > 0$, the prediction error for the estimate
$\fhat_{\PenPar}$ is bounded as
\begin{align}
\| \fhat_{\PenPar} - \fstar\|_\numobs^2 & \leq \frac{2}{\numobs}
\sum_{j=1}^{r} [\KERSHRINK]_{jj}^{2} [U^T \fstar(x_1^\numobs)]_j^2 +
\frac{2}{\numobs} \sum_{j = r+1}^{n}{[U^T \fstar(x_1^\numobs)]_j^2} +
\frac{2}{\numobs} \sum_{j = 1}^{r} \big(1- \KERSHRINK_{jj}\big)^{2}
     [U^T \noise]_j^2.
\end{align}
\elems
\noindent Note that Lemma~\ref{LemRidgeOne} is identical to
Lemma~\ref{LemBiasVarianceBound} with the shrinkage matrices
$\SHRINK{\iter}$ replaced by their analogues $\KERSHRINK$.  See
Appendix~\ref{AppRidgeOne} for the proof of this claim. \\

\noindent Our next step is to show that the diagonal elements of the
shrinkage matrices $\KERSHRINK$ are bounded:
\blems[Properties of kernel ridge shrinkage] 
\label{LemRidgeTwo}
For all indices $j \in
\{1, 2, \ldots, r\}$, the diagonal entries $\KERSHRINK$ satisfy the
bounds
\begin{subequations}
\begin{align}
\label{EqnUpperRidgeBias}	
0 \; \leq \; (\KERSHRINK_{jj})^2 \leq \frac{1}{4 \PenPar
  \kereigemp_j}, \quad \mbox{and} \\
\label{EqnLowerRidgeBias}
\frac{1}{2} \min \big \{ 1, \PenPar \kereigemp_j \big \} \; \leq \;
1-\KERSHRINK_{jj} \; \leq \; \min \big \{1, \PenPar \kereigemp_j \big
\}.
\end{align}
\end{subequations}
\elems
\noindent Note that this is the analog of Lemma~\ref{LemMatrices} from
Theorem~\ref{ThmMain}, albeit with the constant $\frac{1}{4}$ in the
bound~\eqref{EqnUpperRidgeBias} instead of $\frac{1}{2 e}$.  See
Appendix~\ref{AppRidgeTwo} for the proof of this claim.  With these
lemmas in place, the remainder of the proof follows as in the proof of
Theorem~\ref{ThmMain}.


\section{Discussion}

In this paper, we have analyzed the early stopping strategy as applied
to gradient descent on the non-parametric least squares loss.  Our
main contribution was to propose an easily computable and
data-dependent stopping rule, and to provide upper bounds on the
empirical $\LTPn$ error (Theorem~\ref{ThmMain}) and population $\LTP$
error (Theorem~\ref{ThmRandDesign}). We demonstrate in
Corollaries~\ref{CorAchieveSmooth} and ~\ref{CorAchieveFinite} that
our stopping rule yields minimax optimal rates for both low rank
kernel classes and Sobolev spaces. Our simulation results confirm that
our stopping rule yields theoretically optimal rates of convergence
for Lipschitz kernels, and performs favorably in comparison to
stopping rules based on hold-out data and Stein's Unbiased Risk
Estimate. We also showed that early stopping with sum of step-sizes
$\RUNSUM{\iter} = \sum_{k=0}^{t-1}{\Step{k}}$ has a regularization
path that satisfies almost identical mean-squared error bounds as
kernel ridge regression indexed by penalty parameter $\PenPar$.

Our analysis and stopping rule may be improved and extended in a
number of ways. First, it would interesting to see how our stopping
rule can be adapted to mis-specified models. As specified, our method
relies on computation of the eigenvalues of the kernel matrix.  A
stopping rule based on approximate eigenvalue computations, for
instance via some form of sub-sampling~\cite{DrinMah05}, would be
interesting to study as well.


\subsection*{Acknowledgements}
This work was partially supported by NSF grant DMS-1107000 to MJW and BY. In addition, BY was partially supported by the NSF grant SES-0835531 (CDI), ARO-W911NF-11-1-0114 and the Center for Science of
Information (CSoI), an US NSF Science and Technology Center, under grant agreement CCF-0939370, and MJW was also partially supported ONR MURI grant N00014-11-1-086. During this work, GR received partial support from a Berkeley Graduate Fellowship.


\appendix


\section{Derivation of gradient descent updates}
\label{AppGrad}

In this appendix, we provide the details of how the gradient descent
updates~\eqref{EqnGradBvec} are obtained.  In terms of the transformed
vector $\BVEC{} = \EmpKerRoot \, \origweight$, the least-squares
objective takes the form
\begin{align}
\label{EqnLSPar}
\EmpRiskTrans(\BVEC{}) & \defn \frac{1}{2 \numobs} \| \Ydata -
\sqrt{\numobs} \EmpKerRoot \, \BVEC{} \|_2^2 \; = \; \frac{1}{2
  \numobs} \|\Ydata\|_2^2 - \frac{1}{\sqrt{\numobs}}
\inprod{\Ydata}{\EmpKerRoot \, \BVEC{}} + \frac{1}{2} (\BVEC{})^T
\EmpKer \BVEC{}.
\end{align}
Given a sequence $\{\STEP{\iter}\}_{\iter=0}^\infty$, the gradient
descent algorithm operates via the recursion $\BVEC{\iter+1} =
\BVEC{\iter} - \STEP{\iter} \nabla
\EmpRiskTrans(\BVEC{\iter})$. Taking the gradient of $\EmpRiskTrans$
yields
\begin{align*}
\Grad \EmpRiskTrans(\BVEC{}) & = \EmpKer \, \BVEC{} -
\frac{1}{\sqrt{\numobs}} \EmpKerRoot \, \Ydata.
\end{align*}
Substituting into the gradient descent update yields the
claim~\eqref{EqnGradBvec}.

\section{Auxiliary lemmas for Theorem~\ref{ThmMain}}

In this appendix, we collect together the proofs of the lemmas
for Theorem~\ref{ThmMain}.

\subsection{Proof of Lemma~\ref{LemMain}}
\label{AppLemMain}

We prove this lemma by analyzing the gradient descent iteration in an
alternative co-ordinate system.  In particular, given a vector
$\FUN{\iter} \in \real^\numobs$ and the SVD $\EmpKer = U \Dmat U^T$ of
the empirical kernel matrix, we define the vector $\MYPRED{\iter} =
\frac{1} {\sqrt{n}} U^T \FUN{\iter}$.  In this new-coordinate system,
our goal is to estimate the vector $\MYPRED{*} =
\frac{1}{\sqrt{\numobs}} U^T \fstar(x_1^\numobs)$.  Recalling the
alternative form~\eqref{EqnAltGrad} of the gradient recursion, some
simple algebra yields that the sequence $\{\MYPRED{\iter}\}_{\iter
  =0}^\infty$ evolves as
\begin{align*}
\MYPRED{\iter+1} & = \MYPRED{\iter} + \Step{\iter} \widehat{\Lambda}
\frac{\noiseRot}{\sqrt{\numobs}} - \Step{\iter} \widehat{\Lambda}
(\MYPRED{\iter} - \MYPRED{*}),
\end{align*}
where $\noiseRot \defn U^T w$ is a rotated noise vector.  Since
$\MYPRED{0} = 0$, unwrapping this recursion then yields
$\MYPRED{\iter} - \MYPRED{*} = \big (I - \SHRINK{\iter} \big)
\frac{\noiseRot}{\sqrt{\numobs}} - \SHRINK{\iter} \MYPRED{*}$, where
we have made use of the previously defined shrinkage matrices
$\SHRINK{\iter}$.  Using the inequality $\|a+b\|_2^2 \leq 2(\|a\|_2^2
+ \|b\|_2^2)$, we find that
\begin{align*}
\|\MYPRED{\iter} - \MYPRED{*}\|_2^2 & \leq \frac{2}{\numobs} \|(
I-\SHRINK{\iter}) \noiseRot\|_2^2 + 2 \|\SHRINK{\iter}
\MYPRED{*}\|_2^2 \; \stackrel{(i)}{=} \; \frac{2}{\numobs} \|(I-
\SHRINK{\iter}) \noiseRot\|_2^2 + 2 \sum_{j = 1}^{r}
       [\SHRINK{\iter}]_{jj}^2 (\MYPRED{*}_{jj})^2 + 2 \sum_{j =
         r+1}^n (\MYPRED{*}_{jj})^2.
\end{align*}
where step (i) uses the fact that $\kereigemp_j = 0$ for all $j \in
\{r +1, \ldots, \numobs \}$.  Finally, the orthogonality of $U$
implies that $\|\MYPRED{\iter} - \MYPRED{*}\|_2^2 = \frac{1}{n}
\|\FUN{\iter} - \fstar(x_1^\numobs)\|_2^2$, from which the upper
bound~\eqref{EqnMainUpper} follows. \\


\subsection{Proof of Lemma~\ref{LemMatrices}}
\label{AppLemMatrices}

Using the definition of $\SHRINK{\iter}$ and the elementary inequality
$1 - u \leq \exp(-u)$, we have
\begin{align*}
[\SHRINK{\iter}]^2_{jj} = \biggr( \prod_{\tau=0}^{t-1}{(1 -
  \Step{\tau} \kereigemp_j)}\biggr)^2 & \leq \exp(- 2\RUNSUM{\iter}
\kereigemp_j) \; \stackrel{(i)}{\leq} \; \frac{1}{2 e \RUNSUM{\iter}
  \kereigemp_j},
\end{align*}
where inequality (i) follows from the fact that $\sup \limits_{u \in
  \real} \big \{u \exp(-u) \big \} = 1/e$.

Turning to the second set of inequalities, we have $1 -
[\SHRINK{\iter}]_{jj} = 1 - \prod_{\tau=0}^{t-1}{(1 - \Step{\tau}
  \kereigemp_j)}$.  By induction, it can be shown that
\begin{align*} 
1 - [\SHRINK{\iter}]_{jj} & \leq 1 - \max\{0, 1-\RUNSUM{\iter}
\kereigemp_j \} = \min \{1, \RUNSUM{\iter} \kereigemp_j\}.
\end{align*}
As for the remaining claim, we have
\begin{align*}
1 - \prod_{\tau=0}^{t-1}{(1 - \Step{\tau} \kereigemp_j)} &
\stackrel{(i)}{\geq} 1 - \exp(- \RUNSUM{\iter} \widehat{\lambda_i}) \\
& \stackrel{(ii)}{\geq} 1 - (1 + \RUNSUM{\iter}
\widehat{\lambda_i})^{-1} \\
& = \frac{\RUNSUM{\iter} \widehat{\lambda_i}}{1 + \RUNSUM{\iter}
  \widehat{\lambda_i}} \\
& \geq \frac{1}{2}\min \{ 1, \RUNSUM{\iter} \widehat{\lambda_i} \},
\end{align*}
where step (i) follows from the inequality $1- u \leq \exp(-u)$; and
step (ii) follows from the inequality \mbox{$\exp(-u) \leq
  (1+u)^{-1}$,} valid for $u > 0$.


\section{Auxiliary results for Proposition~\ref{PropMainRidge}}
\label{AppRidge}

In this appendix, we prove the auxiliary lemmas used in the proof of
Proposition~\ref{PropMainRidge} on kernel ridge regression.

\subsection{Proof of Lemma~\ref{LemRidgeOne}}
\label{AppRidgeOne}

By definition of the KRR estimate, we have $\biggr(\EmpKer +
\frac{1}{\PenPar}I\biggr)f_{\PenPar}(x_1^\numobs) = \EmpKer
y_1^\numobs$.  Consequently, some straightforward algebra yields the
relation
\begin{align*}
U^T f_\PenPar(x_1^\numobs) & = (I-\KERSHRINK) U^T y_1^\numobs,
\end{align*}
where the shrinkage matrix $\KERSHRINK$ was previously
defined~\eqref{EqnDefnKERSHRINK}. The remainder of the proof follows
using identical steps to the proof of Lemma~\ref{LemMain} with
$\SHRINK{\iter}$ replaced by $\KERSHRINK$.


\subsection{Proof of Lemma~\ref{LemRidgeTwo}}
\label{AppRidgeTwo}

By definition~\eqref{EqnDefnKERSHRINK} of the shrinkage matrix, we
have $[\KERSHRINK]_{jj}^2 = (1 + \PenPar \kereigemp_j)^{-2} \leq
\frac{1}{4 \PenPar \kereigemp_j}$.  Moreover, we also have
\begin{align*}
1-[\KERSHRINK]_{jj} & = 1 - (1 + \PenPar \kereigemp_j)^{-1} =
\frac{\PenPar \kereigemp_j}{1 + \PenPar \kereigemp_j} \leq \min \{1,
\PenPar \kereigemp_j \}, \quad \mbox{and} \\
1 - [\KERSHRINK]_{jj} & = \frac{\PenPar \kereigemp_j}{1 + \PenPar
  \kereigemp_j} \geq \frac{1}{2}\min \{ 1, \PenPar \kereigemp_j\}.
\end{align*}


\section{Properties of the empirical Rademacher complexity}
\label{AppRade}

In this section, we prove that the $\EMPCRIT$ lies in the interval
$(0, \infty)$, and is unique. Recall that the stopping point $\STOP$
is defined as $\EMPCRIT \defn \arg \min \biggr \{\epsilon > 0 \, \mid
\RadEmp_{\EmpKer} \big(\epsilon \big) \leq \epsilon^2/(2 e \sigma)
\biggr \}$.  Re-arranging and substituting for $\RadEmp_{\EmpKer}
\big(\epsilon \big)$ yields the equivalent expression
\begin{equation*}
\EMPCRIT \defn \arg \min \biggr \{ \epsilon > 0 \, \mid
\sum_{i=1}^\numobs \min \big \{ \epsilon^{-2} \kereigemp_i, 1 \big \}
> \numobs \epsilon^2/(4 e^2 \sigma^2) \biggr \} .
\end{equation*}
Note that $\sum_{i=1}^\numobs \min \big \{ \epsilon^{-2} \kereigemp_i,
1 \big \}$ is non-increasing in $\epsilon$ while $n \epsilon^2$ is
increasing in $\epsilon$. Furthermore when $\epsilon = 0$, $ 0 = n
\epsilon^2 < \sum_{i=1}^n \min \big \{ \epsilon^{-2} \kereigemp_i, 1
\big \} > 0$ while for $\epsilon = \infty$, $\sum_{i=1}^n \min \big \{
\RUNSUM{\iter} \kereigemp_i, 1 \big \} < n \epsilon^2$. Hence
$\EMPCRIT$ exists. Further, $\RadEmp_{\EmpKer}(\epsilon)$ is a
continuous function of $\epsilon$ since it is the sum of $n$
continuous functions, Therefore, the critical radius $\EMPCRIT$
exists, is unique and satisfies the fixed point equation
\begin{align*}
\RadEmp_{\EmpKer} \big(\EMPCRIT \big) = \EMPCRIT^2/(2 e \sigma).
\end{align*}

Finally, we show that the integer $\STOP$ belongs to the interval
$[0,\infty)$ and is unique for any valid sequence of step-sizes. Be
  the definition of $\STOP$ given by the stopping
  rule~\eqref{EqnStoppingRule} and $\EMPCRIT$, we have
  $\frac{1}{\RUNSUM{\STOP+1} } \leq \EMPCRIT^2 \leq
  \frac{1}{\RUNSUM{\STOP} }$.  Since $\RUNSUM{0} = 0$ and
  $\RUNSUM{\iter} \rightarrow \infty$ as $t \rightarrow \infty$ and
  $\EMPCRIT \in (0, \infty)$, there exists a unique stopping point
  $\STOP$ in the interval $[0,\infty)$.


\section{Auxiliary results for Theorem~\ref{ThmRandDesign}}

This appendix is devoted to the proofs of auxiliary lemmas used in the
proof for Theorem~\ref{ThmRandDesign}.

\subsection{Proof of Lemma~\ref{LemHilbBound}}
\label{AppLemHilbBound}

Let us write $\FUNIT{\iter} = \sum_{k=0}^{\infty}\sqrt{\lambda_k}
a_{k} \phi_k$, so that $\|\FUNIT{\iter}\|_\Hil^2 = \sum_{k=0}^\infty
a_k^2$.  Recall the linear operator \mbox{$\Phi_X: \ell^2(\Nat)
  \rightarrow \real^\numobs$} defined element-wise via $[\Phi_X]_{jk}
= \phi_j(x_k)$ and the diagonal operator \mbox{$\DiagOpt: \ell^2(\Nat)
  \rightarrow \ell^2(\Nat)$} with entries $[\DiagOpt]_{jj} =
\lambda_j$ and $[\DiagOpt]_{jk} = 0$ for $j \neq k$. By the definition
of the gradient update~\eqref{EqnGradBvec}, we have the relation
\mbox{$a = \frac{1}{\numobs} \DiagOpt^{1/2} \Phi_X^T \EmpKer^{-1}
  f_{\iter}(x_1^\numobs)$.}  Since $\frac{1}{\numobs} \Phi_X \DiagOpt
\Phi_X^T = \EmpKer$,
\begin{align}
\label{EqnCoffee}
\|\FUNIT{\iter}\|_{\Hil}^2 = \|a\|_2^2 = \frac{1}{\numobs}
\FUNIT{\iter}(x_1^\numobs)^T \EmpKer^{-1} \FUNIT{\iter}(x_1^\numobs).
\end{align}
Recall the eigendecomposition $\EmpKer = U \Lambda U^T$ with $\Lambda
= \mbox{diag}(\kereigemp_1, \kereigemp_2, \ldots \kereigemp_r)$, and
the relation $U^T \FUN{\iter} = (I - \SHRINK{\iter}) U^T \Ydata$.
Substituting into equation~\eqref{EqnCoffee} yields
\begin{align*}
\|f_t\|_\Hil^2 & = \frac{1}{n} (\Ydata)^T U (I - \SHRINK{\iter})^2
\Lambda^{-1} U^T \Ydata \\
& \stackrel{(i)}{=} \frac{1}{\numobs} (f^*(x_1^\numobs)+ w)^T U (I -
S_t)^2 \Dmat^{-1} U^T (f^*(x_1^\numobs) + w) \\
& = \underbrace{\frac{2}{\numobs} w^T U (I - S_t)^2 \Dmat^{-1} U^T
  f^*(x_1^\numobs)}_{A_\iter} + \underbrace{\frac{1}{\numobs} w^T U (I
  - S_t)^2 \Dmat^{-1} U^T w}_{B_\iter} + \underbrace{\frac{1}{\numobs}
  f^*(x_1^\numobs)^T U (I - S_t)^2 \Dmat^{-1} U^T
  f^*(x_1^\numobs)}_{C_\iter}
\end{align*}
where equality (i) follows from the observation equation $\Ydata =
\fstar(x_1^\numobs) + w$.  From Lemma~\ref{LemMatrices}, we have
$1-\SHRINK{\iter}_{jj} \leq 1$, and hence $C_\iter \leq
\frac{1}{\numobs} \fstar(x_1^\numobs)^T U \Dmat^{-1} U^T
\fstar(x_1^\numobs) \, \stackrel{(i)}{\leq} \, 1$, where the last step
follows from the analysis in Section~\ref{SecBias}.

\noindent It remains to derive upper bounds on the random variables
$A_t$ and $B_t$.

\paragraph{Bounding $A_t$:}
Since the elements of $w$ are i.i.d, zero-mean and sub-Gaussian with
parameter $\sigma$, we have $\mprob[|A_\iter| \geq 1] \leq 2 \exp(-
\frac{\numobs}{2 \sigma^2 \nu^2})$, where $\nu^2 \defn
\frac{4}{\numobs} [f^*(x_1^\numobs)]^T U (I - S_t)^4
\hat{\Lambda}^{-2} U^T f^*(x_1^\numobs)$.  Since $(1 - (S_\iter)_{jj})
\leq 1$, we have
\begin{align*}
\nu^2 \leq \frac{4}{\numobs} \fstar(x_1^\numobs)^T U (I - S_t)
\Dmat^{-2} U^T f^*(x_1^\numobs) & \leq \frac{4}{n}
\sum_{j=1}^{r}{\frac{[U^T f^*(x_1^\numobs) ]_j^2
  }{\widehat{\lambda}_j^2 } \min(1, \Par_t \widehat{\lambda}_j) } \\
& \leq 4 \, \frac{\Par_t}{\numobs} \sum_{j=1}^{r}{\frac{[U^T
      f^*(x_1^\numobs) ]_j^2 }{\widehat{\lambda}_j } } \\
& \leq 4 \Par_t,
\end{align*}
where the final inequality follows from the analysis in Section~\ref{SecBias}.

\paragraph{Bounding $B_\iter$:}  We begin by  noting that
\begin{align*}
B_t = \frac{1}{\numobs}
\sum_{j=1}^{r}{\frac{(1-S^t_{jj})^2}{\widehat{\lambda_j}} [U^T w]_j^2
} = \frac{1}{\numobs} \trace(U Q U^T, w w^T ),
\end{align*}
where $Q = \diag \{\frac{(1-S^t_{jj})^2}{\widehat{\lambda_j}}, \; j=
1, 2, \ldots r \}$.  Consequently, $B_t$ is a quadratic form in
zero-mean sub-Gaussian variables, and using the tail
bound~\eqref{EqnWrightBound}, we have
\begin{align*}
\mprob \big[ |B_t - \Exs[B_t]| \geq 1 \big[] & \leq \exp(- c \min \{
  \numobs \opnorm{ U QU^T}^{-1} , \numobs^2 \fronorm{U Q U^T}^{-2} \}
  )
\end{align*}
for a universal constant $c$.  It remains to bound $\Exs[B_\iter]$,
$\opnorm{U Q U^T}$ and $\fronorm{U Q U^T}$.

We first bound the mean.  Since $\mathbb{E}[w w^T] \preceq \sigma^2
I_{\numobs \times \numobs}$ by assumption, we have
\begin{align*}
\Exs[B_\iter] \; \leq \frac{\sigma^2}{\numobs} \trace(Q)
\frac{1}{n}\sum_{j=1}^{r} & = \;
(\frac{(1-S^t_{jj})^2}{\widehat{\lambda_j}}) \; \leq
\frac{\Par_t}{n}\sum_{j=1}^{r} \min((\Par_t \widehat{\lambda_j})^{-1},
\Par_t \widehat{\lambda_j})
\end{align*}
But by the definition~\eqref{EqnStoppingRule} of the stopping
rule and the fact that $\iter \leq \STOP$, we have
\begin{align*}
\frac{\Par_t}{n}\sum_{j=1}^{r} \min((\Par_t \widehat{\lambda_j})^{-1},
\Par_t \widehat{\lambda_j}) & \leq \Par_t^2
\Rad^2_{\EmpKer}(1/\sqrt{\RUNSUM{\iter}}) \; \leq \;
\frac{1}{\sigma^2},
\end{align*}
showing that $\Exs[B_\iter] \leq 1$.

Turning to the operator norm, we have
\begin{align*}
\opnorm{U Q U^T} = \max_{j = 1, \ldots, r}
(\frac{(1-S^t_{jj})^2}{\widehat{\lambda_j}}) & \leq \max_{j = 1,
  \ldots, r} \min(\widehat{\lambda_j}^{-1}, \Par_t^2
\widehat{\lambda_j}) \; \leq \Par_t.
\end{align*}
As for the Frobenius norm, we have
\begin{align*}
\frac{1}{n} \fronorm{U Q U^T }^2 = \sum_{j=1}^{r}
(\frac{(1-S^t_{jj})^4}{\widehat{\lambda_j}^2}) & \leq
\frac{1}{\numobs} \sum_{j=1}^{r} \min( \widehat{\lambda_j}^{-2},
\Par_t^4 \widehat{\lambda_j}^2 ) \; \leq
\frac{\Par_t^3}{n}\sum_{j=1}^{r} \min( \Par_t^{-3}
\widehat{\lambda_j}^{-2}, \Par_t \widehat{\lambda_j}^2 ) 
\end{align*}
Using the definition of the empirical kernel complexity, we have
\begin{align*}
\frac{1}{n} \fronorm{U Q U^T }^2 & \leq \Par_t^3
\Rad^2_{\EmpKer}(1/\sqrt{\RUNSUM{\iter}}) \; \leq
\frac{\Par_t}{\sigma^2},
\end{align*}
where the final inequality holds for $\iter \leq {\STOP}$, using the
definition of the stopping rule.

Putting together the pieces, we have shown that 
\begin{align*}
\mprob[ |B_\iter| \geq 2 \quad \mbox{or} \quad |A_\iter| \geq 1] &
\leq \exp(- c \numobs/\Par_\iter)
\end{align*}
 for all $\iter \leq \STOP$.  Since $\frac{1}{\Par_t} \geq \EMPCRIT^2$
 for any $t \leq \STOP$, the claim follows.


\subsection{Proof of Lemma~\ref{LemEmpTru}}
\label{AppLemEmpTru}

In this section, we need to show that $\EMPCRIT \leq \POPCRIT$. Recall
that $\EMPCRIT$ and $\POPCRIT$ satisfy
\begin{equation*}
\RadEmp_{\EmpKer}(\EMPCRIT) = \frac{\EMPCRIT^2}{2 e
  \sigma}\;\;\mbox{and}\;\;\Rad_{\Ker}(\POPCRIT) =
\frac{\POPCRIT^2}{40 \sigma}.
\end{equation*} 
It suffices to prove that $\RadEmp_{\EmpKer}(\POPCRIT) \leq
\frac{\POPCRIT^2}{2 e \sigma}$ using the definition of $\EMPCRIT$.  

In order to prove the claim, we define the random variables
\begin{equation}
\label{EqnDefnRvar}
\RvarHat(w, t) \defn \sup_{\substack{\|g\|_\Hil \leq 1
    \\ \|g\|_\numobs \leq t}} \big| \frac{1}{\numobs}
\sum_{i=1}^\numobs w_i g(x_i) \big|, \quad \mbox{ and } \quad \Rvar(w,
t) \defn \Exs_x \biggr[ \sup_{\substack{\|g\|_\Hil \leq 1 \\ \|g\|_2
      \leq t}} \big| \frac{1}{\numobs} \sum_{i=1}^\numobs w_i g(x_i)
  \big| \biggr],
\end{equation}
where $w_i \sim N(0,1)$ are i.i.d. standard normal, as well as the
associated (deterministic) functions
\begin{align}
\label{EqnDefnRvarExp}
\LocGaussEmp(t) \; \defn \Exs_{w} \big[\RvarHat(w; t) \big] \quad
\mbox{ and } \quad \LocGauss(t) \; \defn \Exs_{w} \big[\Rvar(w; t)
  \big].
\end{align}
By results of Mendelson~\cite{Mendelson02}, there are universal
constants $0 < c_\ell \leq c_u$ such that for all $t^2 \geq
1/\numobs$, we have
\begin{align*}
c_\ell\Rad_{\Ker}(t) \leq \LocGauss(t) \leq c_u \Rad_{\Ker}(t), \quad
\mbox{and} \quad c_\ell \RadEmp_{\EmpKer}(t) \leq \LocGaussEmp(t) \leq
c_u \RadEmp_{\EmpKer}(t).
\end{align*}

We first appeal to the concentration of Lipschitz functions for
Gaussian random variables to show that $\RvarHat(w,t)$ and
$\Rvar(w,t)$ are concentrated around their respective means.  For any
$t > 0$ and vectors $w, w' \in \real^\numobs$, we have
\begin{align*}
|\RvarHat(w,t) - \RvarHat(w',t)| \leq \sup_{\substack{\|g\|_\numobs
    \leq t \\ \|g\|_\Hil \leq 1}} \frac{1}{n}|\sum_{i=1}^{n}(w_i -
w'_i)g(x_i)| \leq \frac{t}{\sqrt{n}} \|w-w'\|_2,
\end{align*}
showing that $w \mapsto \RvarHat(w,t)$ is
$\frac{t}{\sqrt{n}}$-Lipschitz with respect to the $\ell_2$ norm.  A
similar calculation for $w \mapsto \Rvar(w,t)$ shows that
\begin{align*}
|\mathbb{E}_x[\RvarHat(w,t)] - \mathbb{E}_x[\RvarHat(w',t)]| \leq
\mathbb{E}_x[\sup_{\substack{\|g\|_2 \leq t \\ \|g\|_\Hil \leq 1}}
  \frac{1}{n}|\sum_{i=1}^{n}(w_i - w'_i)g(x_i)|] \leq
\frac{t}{\sqrt{n}} \|w-w'\|_2,
\end{align*}
so that it is also Lipschitz $\frac{t}{\sqrt{n}}$.  Consequently,
standard concentration results~\cite{Ledoux01} imply that
\begin{align}
\label{EqnLedOne}
\mprob \big [|\RvarHat(w,t) - \LocGaussEmp(t)| \geq t_0 \big ] \leq 2
\exp\biggr(-\frac{n t_0^2}{2 t^2} \biggr), \quad \mbox{and} \quad
\mprob \big[ |\Rvar(w,t) - \LocGauss(t)| \geq t_0 \big ] \leq 2
\exp\biggr(-\frac{n t_0^2}{2 t^2} \biggr).
\end{align}

Let us condition on the two events $\mathcal{A}(t, t_0) \defn \{
|\RvarHat(w,t) - \LocGaussEmp(t)| \leq t_0 \}$ and
\mbox{$\mathcal{A}'(t, t_0) \defn \{ |\Rvar(w,t) - \LocGauss(t)| \leq
  t_0 \}$.}  We then have
\begin{align*}
\RadEmp_{\EmpKer}(\POPCRIT) \stackrel{(a)}{\leq} \RvarHat(w,\POPCRIT)
+ \frac{\POPCRIT^2}{4e \sigma} \stackrel{(b)}{\leq} \; \Rvar(w,2
\POPCRIT) + \frac{\POPCRIT^2}{4e \sigma} & \stackrel{(c)}{\leq} \; 2
\Rad_{\Ker}(\POPCRIT)+ \frac{3 \POPCRIT^2}{8e \sigma} \;
\stackrel{(d)}{\leq} \; \frac{\POPCRIT^2}{2e \sigma},
\end{align*}
where inequality (a) follows the first bound in
equation~\eqref{EqnLedOne} with $t_0 = \frac{\POPCRIT^2}{4e \sigma}$
and $t = \POPCRIT^2$, inequality (b) follows from
Lemma~\ref{LemGeneral} with $t = \POPCRIT$, inequality (c) follows
from the second bound~\eqref{EqnLedOne} with $t_0 =
\frac{\POPCRIT^2}{8e \sigma}$ and $t = \POPCRIT^2$, and inequality (d)
follows from the definition of $\POPCRIT$.  Since the events
$\mathcal{A}(t, t_0)$ and $\mathcal{A}'(t, t_0)$ hold with the stated
probability, the claim follows.



\bibliographystyle{plain}

\bibliography{May13_Bib}

\end{document}